\documentclass[a4paper,12pt]{article}
\usepackage[utf8]{inputenc}
\usepackage{amsmath,amssymb}
\usepackage[letterpaper]{geometry}
\usepackage{geometry}
\usepackage[inline]{enumitem}
\usepackage{hyperref}
\usepackage[authoryear,semicolon]{natbib}
\setcitestyle{semicolon}
\usepackage{microtype} 
\usepackage{graphicx}  
\usepackage{booktabs}  
\usepackage{multirow}  
\usepackage{graphicx}  
\usepackage{setspace}

\title{A Comparative Study of UMAP and Other Dimensionality Reduction Methods \footnote{The authors gratefully acknowledge the support from NSF DMS-2413833, NIH DE-CTR ACCEL U54-GM104941, and NIH NIGMS DE-INBRE P20 GM103446 for Ding's research, and NSF DMS-2413834, NIH U19 AG068054 ABC-DS, and NIH U19 AG063893 LLF Study for Jin's research. }
	}
 
\author{Guanzhe Zhang$^{\dagger}$, Shanshan Ding$^{\dagger}$, and Zhezhen Jin$^{\ddagger}$}
\date{}

\begin{document}

\doublespacing

\maketitle

\vspace{-.5in}
\begin{center}
    $^{\dagger}$Department of Applied Economics and Statistics, University of Delaware \\
    $^{\ddagger}$Department of Biostatistics, Columbia University 
\end{center}


\begin{abstract}
    Uniform Manifold Approximation and Projection (UMAP) is a widely used manifold learning technique for dimensionality reduction. This paper studies UMAP, supervised UMAP, and several competing dimensionality reduction methods, including Principal Component Analysis (PCA), Kernel PCA, supervised PCA, Sliced Inverse Regression (SIR), Kernel SIR, and t-distributed Stochastic Neighbor Embedding, through a comprehensive comparative analysis. Although UMAP has attracted substantial attention for preserving local and global structures, its supervised extensions, particularly for regression settings, remain rather underexplored. We provide a systematic evaluation of supervised UMAP for both regression and classification using simulated and real datasets, with performance assessed via predictive accuracy on low-dimensional embeddings. Our results show that supervised UMAP performs well for classification but exhibits limitations in effectively incorporating response information for regression, highlighting an important direction for future development.
\end{abstract}

\textbf{Key Words:} dimension reduction; supervised dimension reduction; linear embedding; nonlinear embedding; regression; classification

\section{Introduction}
Dimensionality reduction is crucial in data science and machine learning, especially when working with high-dimensional datasets. It simplifies data for visualization, classification, and other tasks while preserving important structures and reducing complexity \citep{cook2009regression,burges2010dimension,li2018sufficient}. Recent efforts have focused on improving the scalability, clarity, and capacity to manage intricate, non-linear relationships in data. Techniques such as deep learning autoencoders, manifold learning methods, and kernel-based approaches have expanded the range for dimensionality reduction methods, promoting more robust analysis of complex and high-dimensional datasets, particularly in the fields of genomics, image processing, and natural language processing \citep{hinton2006reducing, goodfellow2016deep, becht2019dimensionality}. These methods address challenges where high-dimensional data can lead to overfitting and computational inefficiency \citep{bellman1961adaptive, bishop2006pattern, hastie2009elements, van2009dimensionality}.

In this paper, we study Uniform Manifold Approximation and Projection (UMAP) in both unsupervised and supervised settings, together with several competing dimensionality reduction methods, including PCA, Kernel PCA (KPCA), supervised PCA(SPCA), t-distributed Stochastic Neighbor Embedding (t-SNE), SIR, and kernel SIR (KSIR). While existing studies of supervised UMAP have focused primarily on classification problems, its behavior and effectiveness in regression settings remain largely unexplored. We conduct a systematic empirical evaluation of supervised UMAP for regression and classification, assessing its ability to retain predictive information in the reduced representation and comparing its performance with alternative dimension reduction approaches. Our results demonstrate the strengths and limitations of supervised UMAP for supervised learning tasks and provide guidance for its practical applications.

UMAP, developed by McInnes, Healy, and Melville (2018), is a manifold learning method designed to preserve both local and global structure when reducing high-dimensional data to low-dimensional representations \citep{mcinnes2018umap}. Based on robust topological foundations, it utilizes concepts such as fuzzy simplicial sets and Riemannian geometry to form low-dimensional representations of the original data. Compared with other non-linear dimensionality reduction methods, such as t-SNE, UMAP often provides improved preservation of global structures while maintaining computational efficiency  \citep{becht2019dimensionality}. Subsequent developments have expanded its applicability, including Parametric UMAP, which leverages neural networks to enable scalable embeddings and efficient out-of-sample extension \citep{sainburg2021parametric}. Supervised UMAP, proposed as an extension of the original UMAP framework, incorporates response information into the embedding process to preserve relationships between predictors and outcomes in the reduced low-dimensional space \citep{mcinnes2018umap,umap_api_reference,umap_module_source}. It accommodates both categorical and continuous responses and has been applied in a wide range of supervised learning contexts, particularly in classification settings, including but not limited to genomics, natural language processing, neuroscience, and image analysis \citep{mcinnes2018umap,becht2019dimensionality,pachitariu2019robustness,szubert2019structure,diaz2019umap,kobak2019art,allaoui2020deep}.

Among other competing dimensionality reduction methods, PCA is one of the most well-known linear dimensionality reduction methods 
\citep{pearson1901lines,hotelling1933analysis}. It identifies orthogonal linear combinations of features that maximize variance, thereby preserving information in terms of variability. 
The primary advantages of PCA are its simplicity and interpretability \citep{jolliffe2002principal,cangelosi2007component,ding2014dimension}. However, it assumes that variance is the most important structure in the data, 
and is sensitive to scaling and limited in capturing nonlinear feature relationships \citep{cangelosi2007component, shlens2014tutorial}.  This limitation motivated Kernel PCA, which captures nonlinear structure through kernel mappings \citep{scholkopf1998nonlinear,mika1999kernel}.  PCA and its nonlinear extensions do not incorporate response information when performing dimensionality reduction, which can limit their utility in supervised learning tasks  \citep{burges2010dimension}.

SIR, introduced by Li (1991), is a supervised dimensionality reduction method designed for both regression and classification problems. SIR seeks low-dimensional linear combinations of predictors that capture the essential relationship between the predictors and the response \citep{li1991sliced, cook2009likelihood, cook2009regression,li2018sufficient}. Unlike PCA, which ignores response information, SIR explicitly incorporates the response to identify informative directions for supervised learning. The method assumes that the response depends on a small number of linear combinations of the predictors and operates by partitioning the response into slices and estimating conditional means of the predictors within each slice to recover the central subspace. SIR relies on a linearity condition. This assumption has been shown to be relatively mild in high-dimensional settings \citep{hall1993almost, li2008sliced, qian2019sparse}, and the method has demonstrated strong empirical performance across a wide range of applications, including genomics, econometrics, financial modeling, image analysis, and survival analysis (\cite{ferre1998determining, xia2002adaptive, li2010dimension, ding2015tensor, ding2015higher, jain2017integrative, jain2019sliced, ding2020double}, among others). The idea of the supervised dimension reduction has also been extended to envelope methods and models (\cite{cook2010envelope, cook2013pls, cook2018envelopes, ding2018matrix, wang2018vector, chen2020expectile, ding2021envelope, zhao2022envelopes}).
As with PCA, SIR may be limited in its ability to capture nonlinear relationships. Kernel SIR addresses this limitation by extending SIR to a reproducing kernel Hilbert space to enable the recovery of nonlinear predictor–response relationships while retaining the supervised structure of the original method.

t-SNE, developed by van der Maaten and Hinton (2008), is a widely used non-linear dimensionality reduction and data visualization technique \citep{vandermaaten2008visualizing}. It constructs low-dimensional embeddings by minimizing the divergence between probability distributions of pairwise similarities in the high- and low-dimensional spaces, using Gaussian kernels in the original space and a Student’s t-distribution in the embedding to mitigate the crowding problem \citep{vandermaaten2014accelerating}. t-SNE is particularly effective at preserving local structure and revealing cluster patterns, and has been widely applied to biological data, text embeddings, and image representations \citep{kobak2019art}. Compared with  PCA, t-SNE excels at visualizing nonlinear structure and is well suited for unsupervised exploratory analysis \citep{wattenberg2016how}. However, it does not provide an explicit mapping for new data, requiring recomputation of the embedding when additional observations are introduced \citep{vandermaaten2014accelerating,linderman2019clustering}.

Although UMAP has gained significant attention for its ability to preserve both local and global data structures in low‑dimensional embeddings, research on its supervised extensions remains rather underdeveloped, particularly in regression settings. 
Existing work has focused largely on classification or qualitative visualization, leaving an important knowledge gap in understanding the performance of supervised UMAP for continuous outcomes.  To address this limitation, we investigate supervised UMAP in both classification and regression settings, with a particular emphasis on its performance in regression frameworks. The proposed study provides a comprehensive empirical evaluation of its performance under different modeling conditions and compares it against several widely used dimension reduction and sufficient dimension reduction techniques. To our knowledge, this is the first systematic empirical evaluation of supervised UMAP for both regression and classification problems. It also provides the first comparative study between supervised UMAP and popular sufficient dimension reduction methods, such as SIR and kernel SIR. The study shows that while supervised UMAP performs well in classification, it has limitations to effectively incorporate response information for dimension reduction in regression settings, indicating an important need for improved methods and an important direction for future research.


The rest of the paper is organized in the following way: In section 2, we provide a detailed overview of the methodologies of the dimensionality reduction methods; In section 3, we describe the process of generating the simulated data, including their feature distributions and models to create response variables. We further introduce two real-world datasets used in this study and explain how the evaluation process is structured and performed; Section 4 presents comprehensive comparative numerical results and visualizations, along with a detailed analysis of the findings across methods. Section 5 concludes the paper with a summary of key results and directions for future research.

\section{Methodology}
This section provides the technical background of unsupervised UMAP, supervised UMAP, and other dimensionality reduction methods. Let \(X_1, X_2, \ldots, X_n \in \mathbb{R}^p\) be \(p\)-dimensional predictors and \(Y_1, Y_2, \ldots, Y_n \in \mathbb{R}\) be univariate response variables.

\subsection{UMAP and Supervised UMAP}
UMAP can be developed in both unsupervised and supervised settings. In the unsupervised setting, UMAP utilizes only the feature data to construct embeddings that preserve the global and local structures of the high-dimensional data in a lower-dimensional space \citep{mcinnes2018umap}.

In particular, UMAP builds a fuzzy graph in the following way \citep{mcinnes2018umap}: For each high-dimensional data point \(X_i \in \mathbb{R}^p\), UMAP identifies the \(k\) nearest neighbors and computes distances between \(X_i\) and its neighbors \(X_j\), \(1 \leq j \leq k\), denoted as \(d(X_i, X_j)\) using a distance metric (e.g., Euclidean).

\begin{enumerate}
    \item For each \(X_i\), define \(\rho_i\) and \(\sigma_i\):
          \begin{align}
              \rho_i = \min \left\{ d(X_i, X_j) \mid 1 \leq j \leq k, d(X_i, X_j) > 0 \right\}, \\
              \sum_{j=1}^k \exp \left( -\frac{\max \left( 0, d(X_i, X_j) - \rho_i \right)}{\sigma_i} \right) = \log_2(k),
          \end{align}
          where \(\rho_i\) is the distance to the closest neighbor and \(\sigma_i\) is a local scaling parameter.
    \item Using the two parameters above, construct a similarity graph for data points. For each edge from \(X_i\) to its neighbor \(X_j\), define the weight function as:
          \begin{equation}
              w_{ij} = w(X_i, X_j) = \exp \left( -\frac{\max \left( 0, d(X_i, X_j) - \rho_i \right)}{\sigma_i} \right).
              \label{umap:1}
          \end{equation}
          Symmetrize the weighted graph by modifying the edge weight as:
          \begin{equation}
              \tilde{w}_{ij} = w_{ij} + w_{ji} - w_{ij} w_{ji}.
          \end{equation}
          Here \(\tilde{w}_{ij}\) can be considered as the probability that the given edge exists.
    \item Define the strength of two data points \(z_i,z_j \in \mathbb{R}^d\) in the low-dimensional embedding space as:
          \begin{equation}
              \Phi(z_i, z_j) = \left( 1 + a \left( \| z_i - z_j \|_2^2 \right)^b \right)^{-1},
              \label{eq:umap1}
          \end{equation}
          where $a$ and $b$ are fixed global parameters and are chosen to approximate a target curve $g(s)$:
          \[
              \min_{a,b}
              \sum_{s\in \mathbb{R}_{\ge 0}} 
              \left[
                  \frac{1}{1 + a\, s^{2b}}
                  -
                  g(s)
                  \right]^2
          \]
          with $s= \| z_i - z_j \|_2$,  $g(s)=1$ if $s\le \mathrm{min\_dist}$ and $g(s)=\exp\!\big(-(s - \mathrm{min\_dist}))$ if $s > \mathrm{min\_dist}$. Here $\mathrm{min\_dist}$ is a hyperparameter that defines the minimum  distance between embedded points that are connected.

          The term $\Phi(z_i, z_j) $ defined in \eqref{eq:umap1} can be interpreted as the probability that the two points are connected in the corresponding low-dimensional space, which approaches one as their distance is close to zero and decays in a heavy-tailed pattern as their distance increases. UMAP finds the low-dimensional embeddings \(\{ z_i \in \mathbb{R}^d \}\) by minimizing a cross-entropy loss \(C\) between \(\tilde{w}_{ij}\) and \(\Phi(z_i, z_j)\) using stochastic gradient descent optimization, where
          \begin{equation}
              C = -\sum_{i,j} \left[ \tilde{w}_{ij} \log \left( \Phi(z_i, z_j) \right) + (1 - \tilde{w}_{ij}) \log \left( 1 - \Phi(z_i, z_j) \right) \right].
          \end{equation}
          The initial values of the low-dimensional embeddings in the optimization are obtained through spectral embeddings of the weighted nearest-neighbor graph in the original high-dimensional space \citep{mcinnes2018umap}.

\end{enumerate}

In supervised UMAP, response information is incorporated to enhance the embeddings. How to utilize response information depends on whether the response variable is categorical or continuous, which influences how the response information is integrated into the embedding process. Specifically, supervised UMAP incorporates the information from response variables to modify the construction of the high-dimensional graph and uses it for subsequent computations, while keeping the other steps in the process unchanged. For datasets with categorical response variables, supervised UMAP modifies its similarity graph to pull closer those data points that share the same label and push apart those with different labels. It achieves this by adjusting the edge weights to reward connections between data points with the same label and penalize those with different labels as follows \citep{umap_api_reference,umap_module_source}:
\begin{equation}
    w_{ij}^{sup} = w_{ij}\times
    \begin{cases}
        e^{-f}, & \text{if labels differ}, \\
        1,      & \text{if labels match}.
    \end{cases}
    \label{umap:2}
\end{equation}
Here
\begin{equation}
    f =
    \begin{cases}
        2.5 \times \frac{1}{1-\alpha}, \; & if\;  \alpha < 1 \\
        +\infty, \;                       & if\; \alpha = 1,
    \end{cases}
\end{equation}
where \(\alpha\) is a consistent parameter with a value typically between 0 and 1. After the adjustment, it replaces \(w_{ij}\) with \(w_{ij}^{sup}\)as the edge weight in the similarity graph for the symmetrization and the following process.

For datasets with continuous response variables, the existing supervised UMAP calculates the similarity of numerical response \(Y_i \in \mathbb{R}\) using the same way as it does for computing feature similarity and combines it with feature similarity \citep{umap_api_reference,umap_module_source}. After defining a response distance \(d(Y_i, Y_j)\), the supervised UMAP builds a fuzzy graph for responses in the following way:
\begin{align}
    \rho_{i,y} = \min \left\{ d(Y_i, Y_j) \mid 1 \leq j \leq k, d(Y_i, Y_j) > 0 \right\}, \\
    \sum_{j=1}^k \exp \left( -\frac{\max \left\{ 0, d(Y_i, Y_j) - \rho_{i,y} \right\}}{\sigma_{i,y}} \right) = \log_2(k).
\end{align}
where \(\rho_{i,y}\) is the distance to the closest neighbor in responses and \(\sigma_{i,y}\) is a local scaling parameter. Then compute the response similarity as:
\begin{equation}
    w_{ij,y} = \exp \left( -\frac{\max \left( 0, d(Y_i, Y_j) - \rho_{i,y} \right)}{\sigma_{i,y}} \right).
\end{equation}
Combine it with the edge weight \(w_{ij}\),
\begin{equation}
    w_{ij}^{sup} =
    \begin{cases}
        w_{ij} \cdot (w_{ij,y})^{\frac{\alpha}{1-\alpha}},  & \alpha < 0.5,    \\
        (w_{ij})^{\frac{1- \alpha}{\alpha}} \cdot w_{ij,y}, & \alpha \geq 0.5.
    \end{cases}
    \label{umap:3}
\end{equation}
where \(\alpha\) is a tuning parameter with a value typically between 0 and 1, controlling how strongly the responses impact the overall graph \citep{umap_api_reference,umap_module_source}. Then, supervised UMAP uses \(w_{ij}^{sup}\) as the edge weight in the similarity graph for the symmetrization and the remaining process.

The existing supervised UMAP method for continuous responses may overfit by directly incorporating distance measures based on the response variable. To mitigate this issue, we propose an alternative way that discretizes the response by slicing it into non-overlapping intervals and treating each interval as a categorical label. Supervised UMAP is then applied using edge weights defined in \eqref{umap:2} to construct the embedding. This approach reduces overfitting relative to the existing supervised UMAP method, but it still exhibits limitations for dimensionality reduction in regression settings, as shown in Section 4.

\subsection{PCA, KPCA, and SPCA}
PCA is an unsupervised linear method that reduces dimensionality by projecting the data onto a set of orthogonal components that capture the maximum variance in the dataset \citep{pearson1901lines,hotelling1933analysis,jolliffe2002principal}. It finds the principal components by performing eigenvalue decomposition for the covariance matrix of the original features. Given a dataset of \(n\) samples \(\{X_1, X_2, \ldots, X_n\}\), with each \(X_i \in \mathbb{R}^p\), let’s denote the data matrix as \(X \in \mathbb{R}^{n \times p}\) and center the data matrix as \(X_C\). The covariance matrix \( \Sigma \) is computed as \(\Sigma = \frac{1}{n} X_C^T X_C\). After spectral decomposition on \(\Sigma\), sort eigenvectors in a descending order of their corresponding eigenvalues, and let \(W \in \mathbb{R}^{p \times k}\) be the matrix whose columns are the top \(k\) eigenvectors (principal loadings) of the covariance matrix. Then the projected matrix \(Z \in \mathbb{R}^{n \times k}\) onto \(k\)-dimensional space is given by \(Z = X_C W\), which represents the low dimensional embeddings.

KPCA extends PCA to handle nonlinear data structures by applying kernel functions to project the original data into a higher-dimensional feature space and then performing PCA in that space \citep{scholkopf1998nonlinear,mika1999kernel}. Common kernel functions include Gaussian, polynomial, and sigmoid kernels. KPCA constructs the kernel matrix $K$ for a dataset using a kernel function, where each entry is defined as \(K_{ij} = K(X_i, X_j) = \langle \phi(X_i), \phi(X_j) \rangle\) and  \(\phi\) is a nonlinear mapping that transforms the original data from the input space into a possibly much higher-dimensional feature space, and centers the kernel matrix by \(\tilde{K} = H K H\), where \( H = I_n - \frac{1}{n}\,\mathbf{1}\mathbf{1}^\top \) and
$\mathbf{1} \in \mathbb{R}^n$ is
the vector of all ones \citep{Ma2011KernelPCA}. After spectral decomposition on \( \tilde{K}\), we have the projection of a new data point \(X\) onto the $k$th principal direction as the \(k\)th principal component:
\begin{equation}
    Z_k(X) = \sum_{i=1}^{n} \alpha_{i,(k)}\, \tilde{K}(X_i, X),
\end{equation}
where $\alpha_{(k)} = (\alpha_{1,(k)}, \dots, \alpha_{n,(k)})^{\top}$ is the normalized eigenvector of the centered kernel matrix associated with the $k$-th eigenvalue. KPCA can capture nonlinear relationships in the data, whereas linear methods such as PCA are limited.

To incorporate response information, we use supervised principal components (SPCA) based on the feature-screening approach in \citet{bair2006prediction}.
For each predictor, a univariate score measures its association with
the response in the training data. Predictors with sufficiently large
absolute scores are retained. PCA is then performed on the standardized
training matrix $\widetilde X_S$ containing the retained predictors. If
$W_{S,k}$ contains its $k$ leading right singular vectors, the reduced
representation is
\[
    Z_{\mathrm{SPCA}}=\widetilde X_S W_{S,k}.
\]
Thus, the response guides predictor screening, while the components are
ordinary principal components of the retained predictors. Screening can
reduce the influence of irrelevant variables, particularly when the number
of predictors is large relative to the number of observations
\citep{bair2006prediction,paul2008preconditioning}. A new observation is
embedded using the same retained predictors, training-set means and
standard deviations, and fitted loadings.

\subsection{SIR and KSIR}
SIR is designed for regression and classification tasks and focuses on projecting data onto a lower-dimensional subspace that captures most of the information about the response variable \citep{li1991sliced}. It begins by standardizing the features as $U_i=\hat{\Sigma}^{-\frac{1}{2}}(X_i-\bar X)$, where $\hat{\Sigma}_X$ is the sample covariance matrix of $X$, and slices the range of the response variable $Y$ into \(H\) non-overlapping slices with \(n_h\) observations in the slice $S_h$, $h=1,\dots, H$. It then constructs the covariance of the slice means, weighted by their proportions:
\begin{equation*}
    \hat{V} = \sum_{h=1}^H p_h \hat{m}_h \hat{m}_h^T,
\end{equation*}
where \(p_h\) is the proportion of observations in \(S_h\) and \(\hat{m}_h\) is the sample mean of the features in \(S_h\) \citep{li1991sliced}. SIR performs a spectral decomposition of \( \hat{V}\) and maps the eigenvector back to the original feature space to estimate the effective dimension reduction (EDR) directions. Specifically, if $\hat{v}_k$ is the $k$th eigenvector of $\hat{V}$, the corresponding EDR direction is
$\hat{\beta}_k = \hat{\Sigma}_X^{-\frac{1}{2}} \, \hat{v}_k, \quad k = 1, \dots, d.$
We choose the first few leading EDR directions and project the data onto them to obtain $\hat{Z}_i$, the low-dimensional representation of $X_i$, as:
\begin{equation}
    \hat{Z}_i = \hat{B}^\top (X_i - \bar{X}), \quad \hat{B} = (\hat{\beta}_1, \dots, \hat{\beta}_d).
\end{equation}
The structural dimension $d$ can be selected via information criteria, hypothesis tests, or cross-validation methods.

KSIR extends SIR by incorporating the kernel method similar to KPCA \citep{ScholkopfSmola2002,Wu2008, YehHuangLee2009}. Let \(K_{ij} = K(X_i, X_j) = \langle \phi(X_i), \phi(X_j) \rangle\) map features into a high-dimensional reproducing kernel Hilbert space (RKHS), and let \( K \in \mathbb{R}^{n \times n} \) denote the kernel matrix with entries \( K_{ij} \) with a centered version $\tilde{K} = H K H$,
where \( H = I_n - \frac{1}{n}\,\mathbf{1}\mathbf{1}^\top\), similar to that of KPCA. The response variables $Y_1, \ldots, Y_n$ are divided into $H$ slices.
For each slice $h$, let $\bar{K}_h$ denote the mean of the kernel features within that slice, defined as
$
    \bar{K}_h = \frac{1}{n_h} \sum_{i \in \mathcal{S}_h} \tilde{K}_{\cdot i},
$
where $\mathcal{S}_h$ is the index set of samples in slice $h$, $n_h$ is the number of samples in that slice, and $\tilde{K}_{\cdot i}$ represents the $i$-th column of the centered kernel matrix $\tilde{K}$.
The overall mean of the kernel features is
$
    \bar{K} = \frac{1}{n} \sum_{i=1}^n \tilde{K}_{\cdot i}.
$

KSIR estimates the between-slice covariance matrix in the RKHS as
\[
    M = \sum_{h=1}^{H} p_h (\bar{K}_h - \bar{K})(\bar{K}_h - \bar{K})^\top,
\]
where $p_h = \frac{n_h}{n}$ is the proportion of samples in slice $h$.
The nonlinear effective dimension reduction (EDR) directions are obtained by solving the generalized eigenvalue problem
\[
    \tilde{K}E_H\tilde{K}\alpha = \lambda \tilde{K}\tilde{K} \alpha.
\]
where \(E_H=\sum_{h=1}^{H}\frac{\delta_{h}\,\delta_{h}^{\top}}{n_h}\),  \(\delta_{h}=\big[\delta_h(y_1)\ \cdots\ \delta_h(y_n)\big]^{\top}\), and \(\delta_h(y_i)\) is an indicator function representing if $y_i$ is within slice $h$.
Finally, the projection of a new observation $X$ onto the $k$-th leading KSIR direction is given by
\[
    Z_k(X) = \sum_{i=1}^n \alpha_{i,(k)} K(X_i, X),
\]
where $\alpha_{(k)} = (\alpha_{1,(k)}, \ldots, \alpha_{n,(k)})^\top$ is the $k$-th leading eigenvector.
KSIR captures complex nonlinear relationships between the predictors and the response, which can provide improved performance in cases where traditional SIR fails to uncover nonlinear structures \cite{WuLiangMukherjee2013}.

\subsection{t-SNE}
t-SNE is a nonlinear dimensionality reduction method that aims to preserve local structures in the data, commonly used for visualization in two or three dimensions. It minimizes the divergence between two probability distributions: one representing pairwise similarities in the high-dimensional space and the other in the low-dimensional space\citep{vandermaaten2008visualizing}. In the high-dimensional space, t-SNE applies a Gaussian distribution to model similarities between data points as
\begin{equation}
    p_{j|i} = \frac{\exp \left( -\frac{\| X_i - X_j \|^2}{2 \sigma_i^2} \right)}{\sum_{k \neq i} \exp \left( -\frac{\| X_i - X_k \|^2}{2 \sigma_i^2} \right)},
\end{equation}
and symmetrizes the result with
\begin{equation}
    p_{ij} = \frac{p_{j|i} + p_{i|j}}{2n}.
\end{equation}
In the low-dimensional embedding space, it utilizes a Student t-distribution with one degree of freedom to model similarities between points. Let $Z_i \in \mathbb{R}^d$ and $Z_j \in \mathbb{R}^d$ denote the low-dimensional embeddings corresponding to the high-dimensional data points $X_i$ and $X_j$. The similarity between points $i$ and $j$ in the low-dimensional space is defined as
\begin{equation}
    q_{ij} = \frac{\left( 1 + \| Z_i - Z_j \|^2 \right)^{-1}}{\sum_{k \neq l} \left( 1 + \| Z_k - Z_l \|^2 \right)^{-1}},
\end{equation}
which corresponds to a student’s t-distribution with one degree of freedom, and t-SNE minimizes the Kullback–Leibler divergence between the two distributions:
\begin{equation}
    C = \sum_{i \neq j} p_{ij} \log \frac{p_{ij}}{q_{ij}}
\end{equation}
over $Z_i$, $i=1, \dots,n$, to preserve the similarities of high-dimensional data points as much as possible in the low-dimensional space\citep{vandermaaten2014accelerating}.

\section{Numerical studies}
In this section, we introduce the datasets used in this study and describe the evaluation process. We generate features from three types of distributions and simulate response variables using four different models for each type of feature distribution to create diverse datasets for evaluation. Among the four different models, three produce continuous response variables, while the fourth generates binary outcomes.

\subsection{Simulation settings}
The feature samples $X_i$ ($ i=1,\dots,n$) are generated independently from each of the three distributions:
\begin{enumerate}[label=(\alph*)]
    \item Independent Gaussian features: \(X_i \sim N(0, I_p)\),
    \item Independent Non-Gaussian features: \(X_i \sim (1/2) N(-1_p, I_p) + (1/2) N(1_p, I_p)\),
    \item Correlated Gaussian features: \(X_i \sim N(0, 0.6 I_p + 0.4 1_p 1_p^T)\),
\end{enumerate}
where \(I_p\) denotes an identity matrix of dimension \(p\) and \(1_p\) denotes a \(p\)-dimensional vector with elements all equal to 1.

For the features generated from each of the above, we simulate four distinct types of response variables based on the following models:

\textbf{\\Continuous response variables:}
\begin{itemize*}
    \item Model 1: \(Y_i = \left(\sqrt[3]{\sum_{j=1}^s X_{ij}^3}\right)  \log \left( \sqrt{\sum_{j=1}^s X_{ij}^2} \right) + \epsilon_j\),

    \item Model 2: \(Y_i = \sum_{j=1}^s \frac {X_{ij}}{1 + e^{X_{ij+1}}} + \epsilon_j\),

    \item Model 3: \(Y_i = \sin \left( \frac{\pi \sum_{j=1}^s X_{ij}}{10} \right) + \epsilon_j\),
\end{itemize*}
where \(X_{ij}\) is the $j$th element of \(X_{i}\), \(\epsilon_j\) is the random error, identically and independently distributed as \(N(0, 0.25)\) and independent of \(X_{i}\), and \(s\) is the number of features related to the response variable.

\textbf{\\Categorical response variable:}
\begin{itemize*}
    \item Model 4: \(P(Y_i = 1 \mid X_i) = \frac{1}{1 + \exp \left( -(\beta_0 + \sum_{j=1}^s \beta_j X_{ij}) \right)}\),
\end{itemize*}
where s is the number of features that are related to the probability of generating a binary response, $\beta_0$ is set to $.5$, \( \beta_1, \dots, \beta_s\) are randomly generated from a uniform distribution $\text{Unif}(-2,2)$, and the binary response variables $Y_i$, $i=1, \dots, n$, are generated from a Bernoulli distribution with event probabilities $P(Y_i|X_i)$.

\vspace{.05in}
Therefore, we have a total of 12 simulated data settings. This allows us to evaluate the performance of different dimensionality reduction methods under various data conditions and models. We generate each dataset with a sample size \(n = 1000\), a feature dimension \(p = 500\), and the number of features that are related to the response variable  \(s=10\). We partition each dataset into training data and testing data using a 50-50 ratio, using training data to estimate dimensionality reduction and train predictive models, and then transforming the testing data into embeddings based on the trained models for future effectiveness evaluation. To account for the influence of random variation, we generate 100 datasets for each setting, perform training and testing splits, and estimate the dimensionality reduction using the procedure described above, and then compute the mean and standard error of the predicted results over the 100 datasets for both training and testing sets for final comparisons.

To assess the performance of these dimensionality reduction methods on high-dimensional, low-sample-size(HDLSS) scenarios with continuous responses, we also generate each feature sample with a sample size \(n = 500\), a feature dimension \(p = 8000\), the number of features that are related to the response variable  \(s=10\), and corresponding continuous responses.

\subsection{Real Data}
We apply the public Fashion-MNIST dataset to compare the performance of different dimensionality reduction methods for image classification. The Fashion-MNIST dataset consists of images of fashion items and serves as a more complex alternative to the original MNIST dataset of handwritten digits \citep{xiao2017fashionmnist}. The higher complexity arises from the subtle differences between classes, making it a suitable benchmark for evaluating dimensionality reduction techniques in the context of image data \citep{mukhamediev2024stateoftheart}. The Fashion-MNIST dataset consists of 70,000 grayscale images \citep{tfds2024fashionmnist,keras2025fashionmnist}. Each image has \(28 \times 28\) pixels, representing fashion items. There are 10 classes in the response variable corresponding to different types of clothing and accessories, such as T-shirts, trousers, pullovers, dresses, coats, sandals, shirts, sneakers, bags, and ankle boots. The entire dataset is partitioned into a training dataset with 60,000 images and a test set with 10,000 images \citep{xiao2017fashionmnist}. By applying the dimensionality reduction methods to real-world image data, we evaluate their practical effectiveness and generalization beyond simulated datasets.

The other real data used in this study is the Online News Popularity dataset with a numerical response variable \citep{uci_online_news_popularity}. It contains 39,644 data points extracted from news articles published by Mashable over a two-year period. Each record corresponds to a single article and includes 60 explanatory variables, such as content-based statistics, sentiment and readability measures, keyword metrics, and dummy-coded indicators for publication day and topical category. The response is the number of social media shares an article receives. Because the distribution of the response variable is highly right-skewed, we apply a log transformation before modeling the data. To account for the influence of randomness, we repeat the train–test random sampling 100 times. For each split, we partition the training and testing sets with an equal size, use the training data to estimate dimensionality reduction (or embedding), and then employ it to embed the testing data for further evaluation.

\subsection{Test Data Embedding and Evaluation Procedure}

After fitting each dimensionality reduction method on the training data, test observations are embedded using the fitted representation. For PCA, test observations are centered using the training-set means and projected onto the fitted principal component directions; SPCA uses the same procedure with test data standardized using training-set means and standard deviations. For SIR, test observations are centered using the training-set feature means and projected onto the fitted effective dimension reduction directions, which already incorporate the training-set whitening transformation. KPCA and KSIR compute kernel similarities between test data and training data, apply kernel centering consistently with the training fit, and combine these values with their respective fitted projection coefficients. After fitting UMAP models using the training predictors, test observations were subsequently mapped into the fitted training embedding according to their relationships with nearby training observations. UMAP was not rerun on the combined training and test data. However, t-SNE embeds the training and test predictors jointly and separates the resulting coordinates back into training and test sets.

We employ the K-Nearest Neighbors (KNN) algorithm to evaluate the effectiveness of the reduced representations for subsequent predictions. KNN is a non-parametric learning algorithm that can be used for both classification and regression tasks \citep{CoverHart1967,Altman1992,HTF2009}.
In this study, KNN serves as a consistent evaluation method across different dimensionality reduction techniques. By applying KNN to the training and testing embeddings produced by each method, we assess how well the essential structure and information necessary for accurate predictions are preserved during dimensionality reduction. A higher prediction accuracy rate for classification problems or a lower mean squared error (MSE) for regression problems indicates that the dimensionality reduction method has effectively retained the relevant information from the original data \citep{HTF2009}. We also include ridge regression and random forest as alternative predictive models for simulation data with continuous responses to help determine whether the conclusions are consistent across prediction methods.

\section{Numerical results}
\subsection{Results for simulation data with continuous responses}

To assess the performance of supervised UMAP on datasets with continuous responses, we utilize the response information in three distinct ways during model fitting.
\begin{itemize}
    \item Method 1: Treat it as the original continuous variable, and apply the supervised UMAP with the edge weights defined by \eqref{umap:3}.
    \item Method 2: Treat it as a categorical variable with each unique response value representing a class, and apply the supervised UMAP with edge weights defined by \eqref{umap:2}.
    \item Method 3: Treat it as a categorical variable by slicing the response values into several intervals and treating each interval as a category to replace the original values, and then apply the supervised UMAP with edge weights defined by \eqref{umap:2}.
\end{itemize}

Here, Method 1 and Method 2 are two approaches used in the existing supervised UMAP literature for data with continuous responses, whereas Method 3 is an approach that we design to reduce overfitting from Method 1; we divide the continuous response into five approximately equal-frequency intervals based on its empirical quantiles; specifically, we use the minimum, 20th, 40th, 60th, 80th percentiles, and maximum of the response values as the interval boundaries. In Method 2, when the categories are singletons, each singleton response is assigned equal weight. Because the singleton observation does not share its label with others, it receives no same-category attraction; all feature-graph edges connected to that observation are treated as edges between different categories. As a result, the response information contributes very little to the dimensionality reduction step, and Method 2 behaves almost the same as unsupervised UMAP.

We evaluate the dimensionality reduction methods on the simulated datasets with continuous responses by comparing the MSEs of KNN regression models trained on embeddings of dimensions 2, 3, and 5, using the MSEs of KNN regression models trained on the original datasets as the reference. Tables 1–3 report the results obtained using two-dimensional embeddings for Models 1–3, respectively, under feature distributions (a), (b), and (c). The training MSEs, testing MSEs, and the corresponding standard errors are summarized in the tables. To assess whether the findings are robust across predictive methods, we additionally evaluate ridge regression and random forest on the simulated continuous-response datasets.  The corresponding results are provided in Online Resource 1. The abbreviations used in the tables below are defined as follows:
\begin{itemize}
    \item Orig: Original data without dimension reduction;
    \item SSU: Embeddings from supervised UMAP based on Method 3 (with sliced responses);
    \item CaSU: Embeddings from supervised UMAP based on Method 2;
    \item CoSU: Embeddings from supervised UMAP based on Method 1;
    \item UU: Unsupervised UMAP.
\end{itemize}

These three tables show a consistent pattern among the UMAP variants. CoSU has very low training MSEs but substantially higher testing MSEs than UU, indicating overfitting. SSU improves on CoSU's testing performance but does not outperform UU. CaSU and UU have similar testing MSEs across the simulated settings.

Performance among the other methods depends on the feature distribution. SIR has the lowest testing MSE for a1, b1, a2, b2, and a3. SPCA performs best for b3 and for the correlated-feature datasets c1 and c2, where it also improves on KNN fitted to the original data. For c3, the original data have the lowest testing MSE; SPCA and PCA are the closest reduced-data methods. PCA, KPCA, and UU generally yield similar testing results, while KSIR and t-SNE have particularly large testing MSEs in several mixture and correlated-feature settings. Taken together, the results favor SIR in most independent Gaussian and mixture settings and SPCA in two of the three correlated-feature settings.

\begin{table}[h!]
    \centering
    \caption{Results for Model 1 based on datasets a1, b1, and c1}
    \label{tab:results_a1_b1_c1}
    \resizebox{\textwidth}{!}{
        \begin{tabular}{llccccccccccc}
            \toprule
            Model               & Type     & Orig   & SSU    & CaSU   & CoSU    & UU     & PCA    & KPCA   & SIR    & KSIR    & SPCA   & t-SNE   \\
            \midrule
            \multirow{4}{*}{a1} & TrainMSE & 3.3459 & 2.1683 & 3.7582 & 0.1994  & 3.7164 & 3.8525 & 3.8705 & 1.2472 & 0.1088  & 2.4802 & 3.7320  \\
                                & TrainSE  & 0.0228 & 0.0232 & 0.0294 & 0.0989  & 0.0265 & 0.0305 & 0.0291 & 0.0171 & 0.0020  & 0.0305 & 0.0264  \\
                                & TestMSE  & 6.8812 & 8.5625 & 7.9132 & 9.8923  & 7.9246 & 7.7648 & 7.7688 & 5.6527 & 11.5459 & 6.2568 & 7.9251  \\
                                & TestSE   & 0.0426 & 0.0597 & 0.0492 & 0.1286  & 0.0558 & 0.0482 & 0.0485 & 0.0391 & 0.0610  & 0.0490 & 0.0382  \\
            \midrule
            \multirow{4}{*}{b1} & TrainMSE & 1.8150 & 0.9296 & 2.0119 & 0.1731  & 2.0029 & 2.0205 & 2.0525 & 0.9621 & 0.3177  & 1.3990 & 1.9781  \\
                                & TrainSE  & 0.0175 & 0.0107 & 0.0165 & 0.0676  & 0.0168 & 0.0173 & 0.0178 & 0.0141 & 0.0076  & 0.0164 & 0.0177  \\
                                & TestMSE  & 3.7325 & 4.7323 & 4.1784 & 10.0831 & 4.1348 & 4.0642 & 4.1327 & 3.1404 & 54.1738 & 3.3084 & 46.5391 \\
                                & TestSE   & 0.0261 & 0.0422 & 0.0328 & 0.6822  & 0.0313 & 0.0307 & 0.0326 & 0.0288 & 0.2597  & 0.0383 & 4.3907  \\
            \midrule
            \multirow{4}{*}{c1} & TrainMSE & 1.2051 & 0.8454 & 1.3014 & 0.1356  & 1.3078 & 1.2686 & 1.2840 & 1.1590 & 0.3106  & 0.8835 & 1.2499  \\
                                & TrainSE  & 0.0105 & 0.0092 & 0.0109 & 0.0438  & 0.0114 & 0.0101 & 0.0121 & 0.0118 & 0.0071  & 0.0104 & 0.0116  \\
                                & TestMSE  & 2.4139 & 2.9321 & 2.7095 & 4.3873  & 2.6795 & 2.5472 & 2.5896 & 2.5951 & 17.5376 & 2.0651 & 10.1491 \\
                                & TestSE   & 0.0170 & 0.0214 & 0.0205 & 0.1822  & 0.0188 & 0.0189 & 0.0215 & 0.0193 & 0.1302  & 0.0200 & 0.9223  \\
            \bottomrule
        \end{tabular}
    }
\end{table}

Intuitively, supervised methods are expected to outperform unsupervised approaches because they explicitly leverage information from the response variable. However, the supervised UMAP approach CoSU yields among the highest testing MSEs and performs worse than its unsupervised counterpart, suggesting that response information is not effectively incorporated into the current supervised UMAP framework for dimension reduction.  In contrast, the linear supervised method SIR consistently achieves relatively small MSEs. This shows that SIR is able to more effectively exploit response information and capture the underlying predictor–response relationships.

\begin{table}[h!]
    \centering
    \caption{Results for Model 2 based on datasets a2, b2, and c2}
    \label{tab:results_a2_b2_c3}
    \resizebox{\textwidth}{!}{
        \begin{tabular}{llccccccccccc}
            \toprule
            Model               & Type     & Orig   & SSU    & CaSU   & CoSU    & UU     & PCA    & KPCA   & SIR    & KSIR    & SPCA   & t-SNE   \\
            \midrule
            \multirow{4}{*}{a2} & TrainMSE & 1.8710 & 1.1763 & 2.0956 & 0.1249  & 2.0753 & 2.1044 & 2.1242 & 0.5773 & 0.1098  & 1.1764 & 2.0647  \\
                                & TrainSE  & 0.0166 & 0.0129 & 0.0176 & 0.0595  & 0.0177 & 0.0193 & 0.0177 & 0.0057 & 0.0026  & 0.0126 & 0.0186  \\
                                & TestMSE  & 3.6844 & 4.6437 & 4.2859 & 5.2758  & 4.2849 & 4.2955 & 4.2922 & 2.5370 & 6.3686  & 3.0519 & 4.3799  \\
                                & TestSE   & 0.0281 & 0.0334 & 0.0386 & 0.0900  & 0.0321 & 0.0307 & 0.0374 & 0.0208 & 0.0404  & 0.0278 & 0.0323  \\
            \midrule
            \multirow{4}{*}{b2} & TrainMSE & 2.5760 & 1.2935 & 2.9138 & 0.2829  & 2.8395 & 2.8876 & 2.9560 & 1.0956 & 0.3750  & 1.6549 & 2.8810  \\
                                & TrainSE  & 0.0237 & 0.0184 & 0.0289 & 0.0494  & 0.0293 & 0.0309 & 0.0287 & 0.0189 & 0.0087  & 0.0229 & 0.0276  \\
                                & TestMSE  & 5.0703 & 6.7569 & 5.8695 & 13.6292 & 5.9508 & 5.7387 & 5.9544 & 3.9864 & 58.3359 & 4.2230 & 55.5305 \\
                                & TestSE   & 0.0521 & 0.0760 & 0.0634 & 0.8127  & 0.0634 & 0.0571 & 0.0568 & 0.0469 & 0.2939  & 0.0626 & 4.6483  \\
            \midrule
            \multirow{4}{*}{c2} & TrainMSE & 1.4017 & 0.9936 & 1.5214 & 0.2012  & 1.5371 & 1.4587 & 1.4810 & 1.3240 & 0.5932  & 0.9304 & 1.4414  \\
                                & TrainSE  & 0.0134 & 0.0128 & 0.0156 & 0.0443  & 0.0153 & 0.0140 & 0.0155 & 0.0147 & 0.0178  & 0.0148 & 0.0143  \\
                                & TestMSE  & 2.8682 & 3.4605 & 3.2473 & 5.4376  & 3.2465 & 3.0055 & 3.1030 & 3.0885 & 26.1289 & 2.3360 & 18.8163 \\
                                & TestSE   & 0.0265 & 0.0367 & 0.0319 & 0.2680  & 0.0295 & 0.0246 & 0.0360 & 0.0279 & 0.1818  & 0.0336 & 1.3457  \\
            \bottomrule
        \end{tabular}
    }
\end{table}

Both supervised UMAP and SIR use response information for dimensionality reduction, but they do so in fundamentally different ways. In supervised UMAP, response information is used to reweigh the feature fuzzy graph, encouraging observations that are similar in both predictor and response spaces to remain close in the embedding. However, it does not explicitly model the relationship between X and Y. In contrast, SIR estimates dimension-reduction directions by exploiting changes in the conditional distribution of X given Y, making its objective more directly related to identifying predictor directions that carry information about the response. Therefore, supervised UMAP is best understood as a label-informed dimensionality-reduction method: although its embedding may enhance downstream prediction, its optimization objective does not directly guarantee improved predictive performance.

\begin{table}[h!]
    \centering
    \caption{Results for Model 3 based on datasets a3, b3, and c3}
    \label{tab:results_a3_b3_c3}
    \resizebox{\textwidth}{!}{
        \begin{tabular}{llccccccccccc}
            \toprule
            Model               & Type     & Orig   & SSU    & CaSU   & CoSU   & UU     & PCA    & KPCA   & SIR    & KSIR   & SPCA   & t-SNE  \\
            \midrule
            \multirow{4}{*}{a3} & TrainMSE & 0.4262 & 0.2548 & 0.4568 & 0.0421 & 0.4537 & 0.4583 & 0.4584 & 0.1629 & 0.0134 & 0.2958 & 0.4548 \\
                                & TrainSE  & 0.0032 & 0.0030 & 0.0035 & 0.0102 & 0.0034 & 0.0036 & 0.0039 & 0.0017 & 0.0002 & 0.0030 & 0.0034 \\
                                & TestMSE  & 0.8458 & 1.0307 & 0.9301 & 1.2168 & 0.9243 & 0.9283 & 0.9315 & 0.6927 & 1.3777 & 0.7613 & 0.9384 \\
                                & TestSE   & 0.0047 & 0.0074 & 0.0069 & 0.0203 & 0.0069 & 0.0060 & 0.0075 & 0.0039 & 0.0071 & 0.0061 & 0.0052 \\
            \midrule
            \multirow{4}{*}{b3} & TrainMSE & 0.4345 & 0.2645 & 0.4593 & 0.0532 & 0.4529 & 0.4687 & 0.4668 & 0.1946 & 0.0139 & 0.2959 & 0.4541 \\
                                & TrainSE  & 0.0037 & 0.0028 & 0.0034 & 0.0087 & 0.0036 & 0.0033 & 0.0035 & 0.0028 & 0.0002 & 0.0067 & 0.0036 \\
                                & TestMSE  & 0.8587 & 1.0295 & 0.9298 & 1.2990 & 0.9236 & 0.9167 & 0.9230 & 0.7074 & 1.3890 & 0.6925 & 0.9463 \\
                                & TestSE   & 0.0049 & 0.0064 & 0.0068 & 0.0244 & 0.0058 & 0.0049 & 0.0060 & 0.0050 & 0.0078 & 0.0123 & 0.0150 \\
            \midrule
            \multirow{4}{*}{c3} & TrainMSE & 0.3142 & 0.1971 & 0.3283 & 0.0541 & 0.3282 & 0.3191 & 0.3223 & 0.3040 & 0.0131 & 0.3122 & 0.3173 \\
                                & TrainSE  & 0.0026 & 0.0025 & 0.0027 & 0.0100 & 0.0024 & 0.0023 & 0.0025 & 0.0029 & 0.0002 & 0.0027 & 0.0028 \\
                                & TestMSE  & 0.6308 & 0.7446 & 0.6650 & 0.8559 & 0.6667 & 0.6415 & 0.6497 & 0.6596 & 1.5164 & 0.6413 & 0.9659 \\
                                & TestSE   & 0.0040 & 0.0055 & 0.0040 & 0.0171 & 0.0039 & 0.0038 & 0.0039 & 0.0040 & 0.0083 & 0.0046 & 0.0346 \\
            \bottomrule
        \end{tabular}
    }
\end{table}

Both Continuous-response supervised UMAP (CoSU) and unsupervised UMAP (UU) construct a fuzzy graph for the data points, but Continuous-response supervised UMAP is subject to a neighborhood-overlap bottleneck.  As shown in \eqref{umap:3}, supervised UMAP  CoSU constructs a fuzzy graph from the continuous response and combines it with the predictor graph. It evaluates distances in the predictor and response spaces independently prior to fusion, which can force artificial synthesis of conflicting manifolds. It can use the response effectively only when its local neighborhoods overlap sufficiently with the predictor-space neighborhoods. If an edge appears in one graph but not the other, UMAP will set a very small value for the combined edge. Low overlap may occur in many situations, such as when the data contain a large number of irrelevant predictors, the responses are noisy, or the regression function maps distinct feature vectors to identical or similar response values. In that case, the response graph cannot reliably introduce response-relevant relationships that are absent from the predictor graph, and the resulting supervision may be less informative, unstable, or difficult to transfer to test observations. This may explain why unsupervised UMAP outperforms supervised UMAP in most of our numerical studies involving data with a continuous response.

To assess whether performance is consistent across low-dimensional spaces, we evaluate these methods using embeddings of dimensions 2, 3, and 5. The results show that the overall performance pattern is largely stable. The best-performing method remains unchanged across all nine simulation settings when the dimension increases from 2 to 3. At dimension 5, the only changes occur for b1 and b2: SIR performs best at dimensions 2 and 3, whereas SPCA performs best at dimension 5. The UMAP variants also show stable performance, with CoSU consistently yielding the highest test MSE. Detailed results for the 3- and 5-dimensional embeddings are provided in Online Resource 1.

The results from the additional predictive models, ridge regression and random forest, presented in Online Resource 1, further support the consistency of our main conclusions. In particular, unsupervised UMAP outperforms continuous-response supervised UMAP in nearly all settings. Thus, incorporating response information into UMAP does not yield a consistent predictive advantage, regardless of whether the downstream model relies on local neighborhoods, linear relationships, or nonlinear tree-based prediction. Across the three predictive models and nine datasets, SIR and SPCA generally perform best among the dimensionality-reduction methods, yielding the best reduced representation in 25 of the 27 model-dataset combinations.

In the high-dimensional, low-sample-size (HDLSS) scenarios with 500 observations and 8000 predictors, the performance pattern among the UMAP variants remains broadly consistent with that in the original simulation setting: CaSU and unsupervised UMAP generally perform similarly, SSU is usually less accurate, and CoSU continues to exhibit a large training-test gap. The relative performance of SIR and SPCA changes substantially. In the original setting, SIR achieves the lowest test MSE in five of the nine scenarios, whereas SPCA is best in three. Under the HDLSS setting, SPCA achieves the
lowest test MSE in seven scenarios. The corresponding results are provided in Online Resource 1. SPCA performs better than SIR in this experiment because its supervised feature- screening step is well suited to the sparse simulation design, in which only 10 of the 8,000 predictors are related to the response. SPCA first selects predictors according to their association with the response and then performs PCA on the retained subset, substantially reducing the number of irrelevant variables entering the component estimation. In contrast, the conventional SIR implementation estimates inverse-regression directions using all predictors and requires covariance whitening. When the number of predictors greatly exceeds the number of training observations, the sample covariance matrix is singular, and the estimated inverse-regression directions become highly
sensitive to sampling noise.

\subsection{Evaluation of continuous-response supervised UMAP embeddings}

To further examine the local structure of the test embeddings produced by continuous-response supervised UMAP, we use the following measures to assess neighborhood preservation relative to the original feature space and response similarity within embedding-space neighborhoods.

\subsubsection{Test Data Neighborhood Preservation}

Let \(N_{X,k}(i)\) denote the set of the \(k\)
nearest test observations to observation \(i\) in the original
feature space. Similarly, let \(N_{Z,k}(i)\) denote the
corresponding set of nearest test observations in the UMAP embedding.
Observation \(i\) itself is excluded from both sets.

For each test observation, calculate the proportion of its
original-space neighbors that are retained in the embedding:

\[O_{i}(k) = \frac{\left| N_{X,k}(i) \cap N_{Z,k}(i) \right|}{k}\]

Average this quantity over all test observations:

\[O(k) = \frac{1}{n_{test}}\sum_{i = 1}^{n_{test}}{O_{i}(k)}\]

{The interpretation of $O(k)$ is:}

If  \(O(k)=1\): all original-space neighbors are preserved in the embedding.

If  \(O(k)=0\): none of the original-space neighbors are preserved.

Larger values indicate better preservation of local neighborhood
geometry.






We perform the supervised continuous-response UMAP on the Data\_a1
dataset using the same train-test split and setting the number of
neighbors $k$ to  be 15. The overlap measure  is \(O(15) = 0.0329\), The proportion of overlapping neighborhoods in  test dataset is 3.29\% on average, indicating small overlapping between the embedding and original feature spaces.

\subsubsection{Response Similarity within Test Neighborhoods}

This algorithm evaluates whether observations that are close in the test
embedding have similar responses. It measures the variance of all
responses in each neighborhood. Smaller variance indicates greater local
response similarity.

Let $n$ be the number of test observations, \(Z_{i}\) the embedding of
test observation \(i\), and \(Y_{i}\) its observed response. Choose an
integer neighborhood size \(k\) with \(1 \leq k < n\). Define
\(S_{Z,k}(i)\) as the neighborhood that contains observation \(i\) and
its \(k\) nearest neighbors, and \(s_{Y,test}^{2}\) as the sample
variance of the entire test dataset.

For each test observation \(i\), find its \(k\) nearest neighbors using
Euclidean distance in the embedding space. Calculate the neighborhood
response mean and the total squared variation within the neighborhood as:
${\overline{Y}}_{Z,i} = \frac{1}{k + 1}\sum_{j \in S_{Z,k}\left( i \right)}^{}Y_{j}$ and
${SS}_{Z,i} = \sum_{j \in S_{Z,k}\left( i \right)}^{}\left( Y_{j} - {\overline{Y}}_{Z,i} \right)^{2}.$ The smaller the ${SS}_{Z,i}$, the more similar the response $i$ neighborhood.

Average within and across neighborhoods:
\[V_{Z}\left( k \right) = \frac{1}{n}\sum_{i = 1}^{n}\frac{{SS}_{Z,i}}{k } = \frac{1}{n k }\sum_{i = 1}^{n}{SS}_{Z,i}.\]

And normalize it based on the total sample variance of the test dataset:
\[V_{Z}^{*}\left( k \right) = \frac{V_{Z}\left( k \right)}{s_{Y,test}^{2}}. \] 

Repeat the above steps in the original feature space to obtain
response neighborhood  measure \(V_{X}^{*}\left( k \right)\), and compare the two
measures based on the embedding space and the original feature space by:

\[R_{V}\left( k \right) = \frac{V_{Z}^{*}\left( k \right)}{V_{X}^{*}\left( k \right)}.\]

The interpretation of \(R_{V}(k)\) is:

If   \(R_{V}\left( k \right) < 1\): embedding neighborhoods have lower
average response dispersion.

If    \(R_{V}\left( k \right) \approx 1\): average response dispersion is
comparable.

If    \(R_{V}\left( k \right) > 1\): embedding neighborhoods have higher
average response dispersion.

We evaluated this algorithm using a continuous-response supervised UMAP
model on the Data\_a1 dataset, with the number of neighbors $k$ set to  be 15.
The embedding-space test response similarity measure $V_{Z}^{*}\left( k \right)$  is 0.9196; the
original-space test response similarity $V_{X}^{*}\left( k \right)$ is 0.8271, yielding a ratio
\(R_{V}(15)\) of 1.1174. Thus, the embedding-space neighborhoods exhibit
slightly greater response dispersion than the corresponding
neighborhoods in the original space.

The results indicate that continuous-response supervised UMAP preserves little of the original test-data neighborhood structure and does not improve local response similarity,  suggesting limited effectiveness in organizing test observations according to their responses in this setting.

\subsection{Results for simulation data with categorical responses}

The evaluation in this section is conducted on datasets with categorical responses, where the response variables are generated according to Model 4 and the predictors are drawn from distributions (a), (b), and (c). The datasets are named as a4, b4, and c4 in Table 4. To assess classification performance, we compare misclassification error rates obtained from KNN classification applied to the embedded test datasets as well as to the original test data. The corresponding training error rates (TrainError), testing error rates (TestError), and the standard errors of the testing error rates, computed from 100 random repetitions, are reported in Table 4.

Because the response variable is categorical, the CaSU and SSU approaches coincide and therefore produce identical results, while CoSU is not applicable in this setting. Accordingly, CoSU and SSU are not reported in Table 4.

\begin{table}[h!]
    \centering
    \caption{Results (misclassification rate) for Model 4 based on datasets a4, b4, and c4}
    \begin{tabular}{llcccccccc}
        \toprule
        Model    & Type       & Orig  & CaSU  & UU    & PCA   & KPCA  & SIR   & KSIR  & t-SNE \\
        \midrule
        Data\_a4 & TrainError & 0.223 & 0.033 & 0.243 & 0.246 & 0.246 & 0.087 & 0.013 & 0.242 \\
                 & TestError  & 0.451 & 0.447 & 0.490 & 0.493 & 0.495 & 0.357 & 0.496 & 0.495 \\
                 & TestSE     & 0.002 & 0.002 & 0.002 & 0.003 & 0.003 & 0.003 & 0.002 & 0.002 \\
        \midrule
        Data\_b4 & TrainError & 0.156 & 0.039 & 0.167 & 0.170 & 0.172 & 0.141 & 0.011 & 0.164 \\
                 & TestError  & 0.286 & 0.280 & 0.304 & 0.304 & 0.307 & 0.284 & 0.495 & 0.510 \\
                 & TestSE     & 0.015 & 0.014 & 0.015 & 0.015 & 0.016 & 0.013 & 0.002 & 0.023 \\
        \midrule
        Data\_c4 & TrainError & 0.197 & 0.039 & 0.206 & 0.205 & 0.207 & 0.173 & 0.012 & 0.199 \\
                 & TestError  & 0.370 & 0.358 & 0.387 & 0.392 & 0.392 & 0.370 & 0.493 & 0.461 \\
                 & TestSE     & 0.008 & 0.008 & 0.008 & 0.009 & 0.008 & 0.007 & 0.002 & 0.010 \\
        \bottomrule
    \end{tabular}
\end{table}

As shown in Table 4, the supervised UMAP method CaSU and SIR outperform the other dimensionality reduction approaches on the test datasets. In this classification setting, supervised UMAP exhibits strong performance and achieves competitive classification accuracy. These results suggest that, in contrast to the regression settings, response information is more effectively incorporated in the supervised UMAP framework when the response is categorical. Correspondingly, supervised UMAP provides clear improvements over its unsupervised method in terms of predictive performance. These findings indicate that, when appropriately leveraged, response information in the supervised UMAP framework can enhance prediction accuracy for supervised learning tasks. 

\subsection{Results for real data with categorical responses}

The first real data used in our application is the Fashion‑MNIST dataset described in Section 3.2. Table 5 reports the misclassification error rate for each method on both the training and testing datasets, along with the elapsed time (in seconds) for model fitting, data transformation, and evaluation.

\begin{table}[h!]
    \centering
    \caption{Results (misclassification rate)  for Fashion-MNIST dataset}
    \begin{tabular}{lccc}
        \toprule
        Type              & TrainError & TestError & Elapsed Time (sec) \\
        \midrule
        Original          & 0.080      & 0.146     & 40.524             \\
        UMAP Supervised   & 0.018      & 0.162     & 18.909             \\
        UMAP Unsupervised & 0.145      & 0.247     & 13.922             \\
        PCA               & 0.305      & 0.517     & 1.301              \\
        KPCA              & 0.305      & 0.515     & 35.173             \\
        SIR               & 0.292      & 0.498     & 1.125              \\
        KSIR              & 0.470      & 0.762     & 8.474              \\
        t-SNE             & 0.092      & 0.504     & 220.947            \\
        \bottomrule
    \end{tabular}
\end{table}

The original data achieves balance between training and testing performance without dimensionality reduction and serves as the baseline for comparison. Supervised UMAP delivers the strongest training results, and its testing misclassification rate remains close to that of the original data after compressing the 784‑dimensional features into two‑dimensional embeddings. Figure 1 displays the reduced two‑dimensional embeddings for both the training and testing sets, colored by fashion‑item class. The visualization shows that supervised UMAP cleanly separates the training data into the expected classes while preserving the global structure. The testing data aligns well with this structure, exhibiting clear class separation and producing the highest prediction accuracy. For consistency across comparisons and for ease of visualization, we set the embedding dimension to 2. Results obtained with higher dimensional embeddings, or with embeddings obtained by cross-validation show a similar pattern.

In contrast, unsupervised UMAP shows reduced performance on both the training and testing sets, and the corresponding scatter plots reveal a noticeable decline in embedding quality. For both training and testing datasets, class boundaries are less distinct than those in the supervised embeddings, and some regions, particularly in the lower‑right portion of the plot, show substantial mixing of classes.

\begin{figure}[h!]
    \centering
    \includegraphics[width=0.8\textwidth]{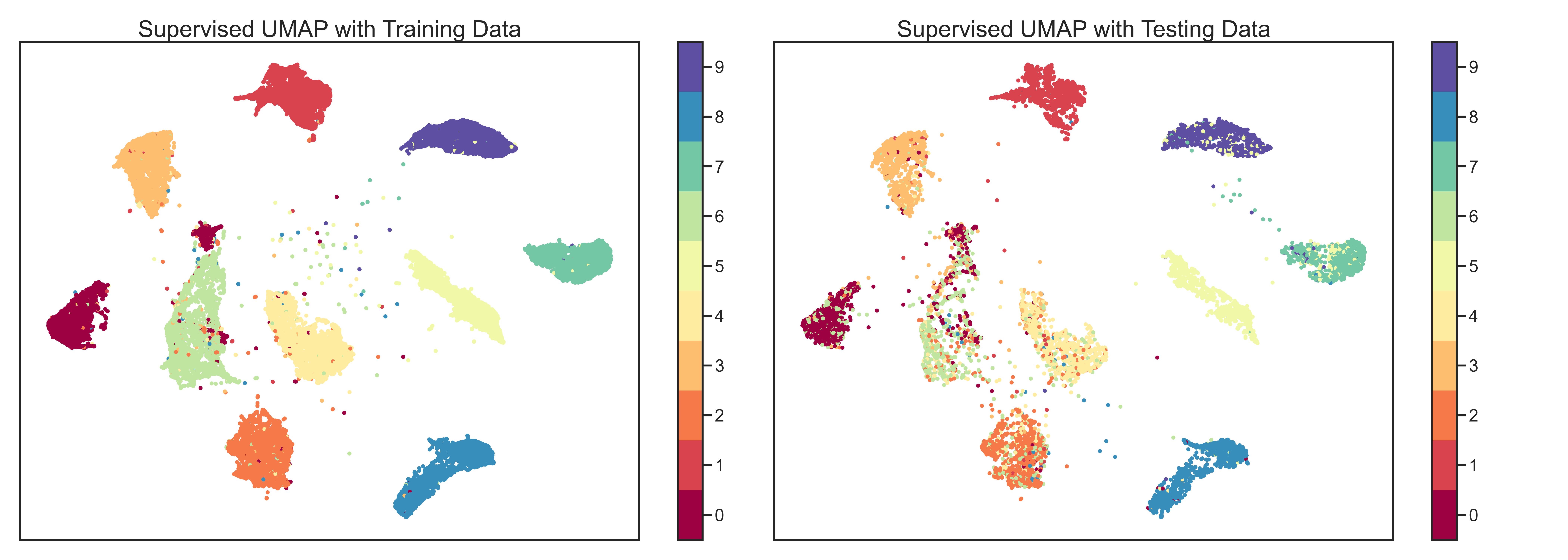}
    \includegraphics[width=0.8\textwidth]{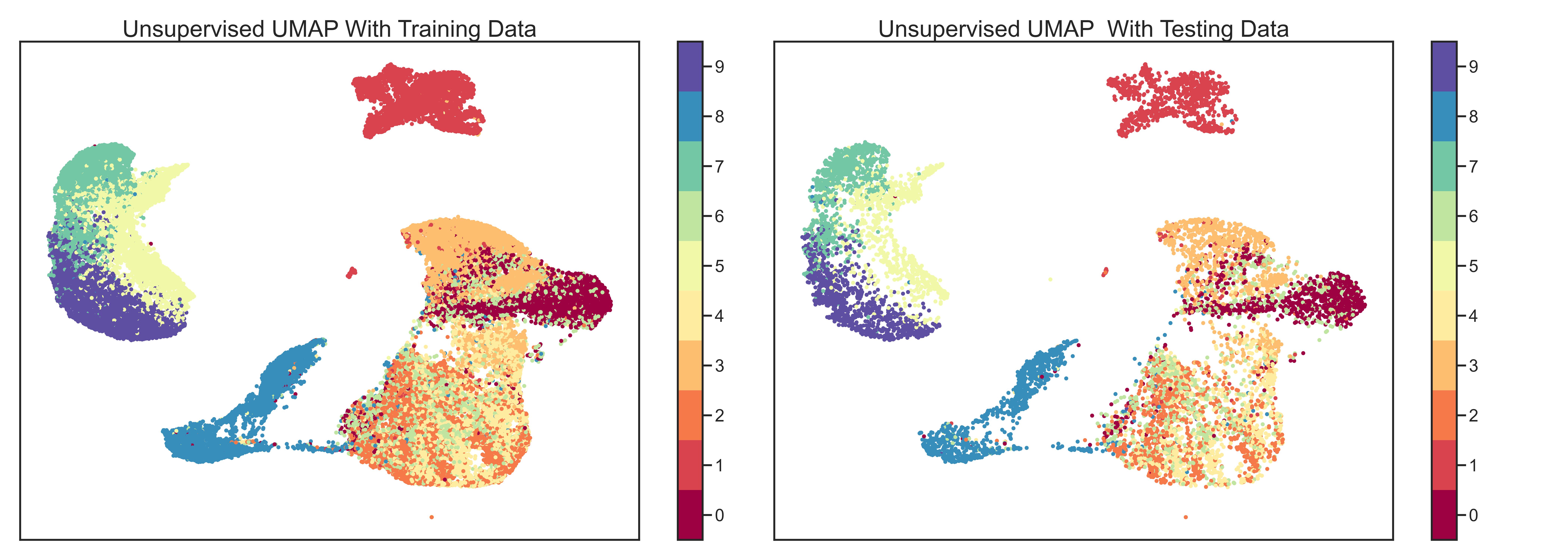}
    \caption{Performance plots of UMAP}
    \label{fig:UMAP} 
\end{figure}

PCA and SIR are fast but less effective for this large-scale dataset. There show higher misclassification error rates than the UMAP methods. Both PCA and SIR are linear embedding methods, focusing on maximizing variance  or preserving sufficient effective reductions using linear transformations. However, the Fashion-MNIST dataset has complex, non-linear embedding patterns that PCA and SIR might not be able to capture. Neither KPCA nor KSIR performs better than their non-kernel counterparts on this dataset. Similar to the misclassification rates, their embedding plots show less clear boundaries and less effective embeddings for classification.

\begin{figure}[h!]
    \centering
    \includegraphics[width=0.8\textwidth]{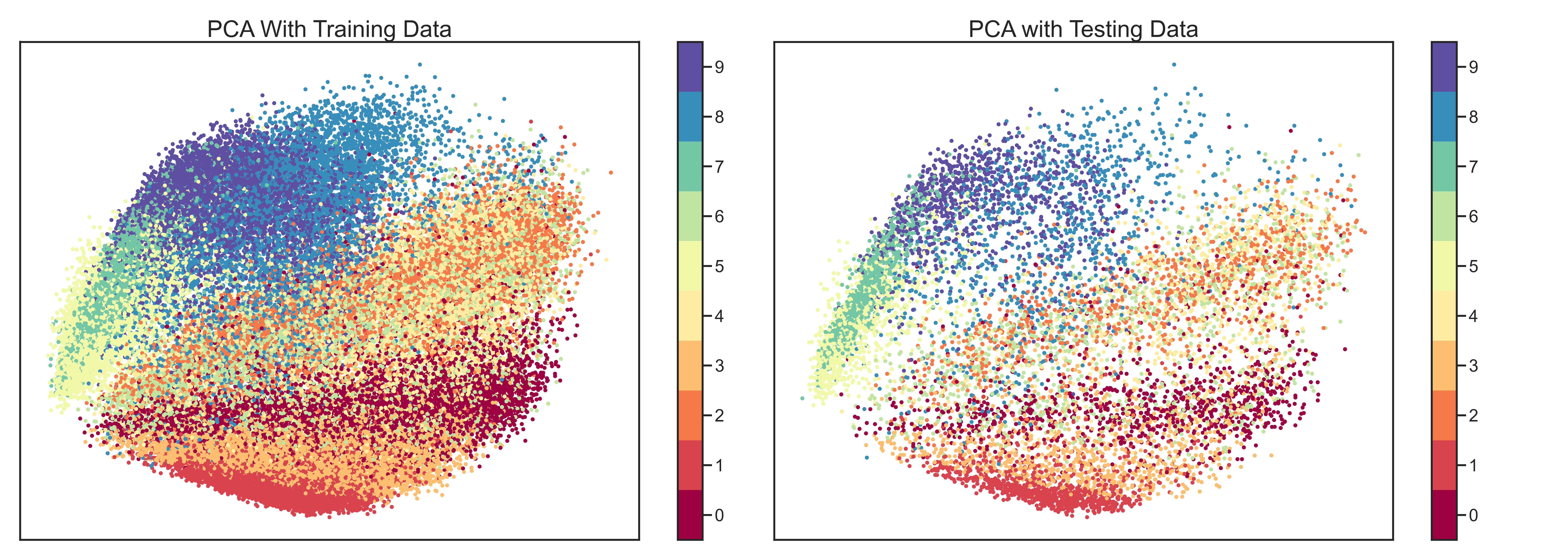}
    \includegraphics[width=0.8\textwidth]{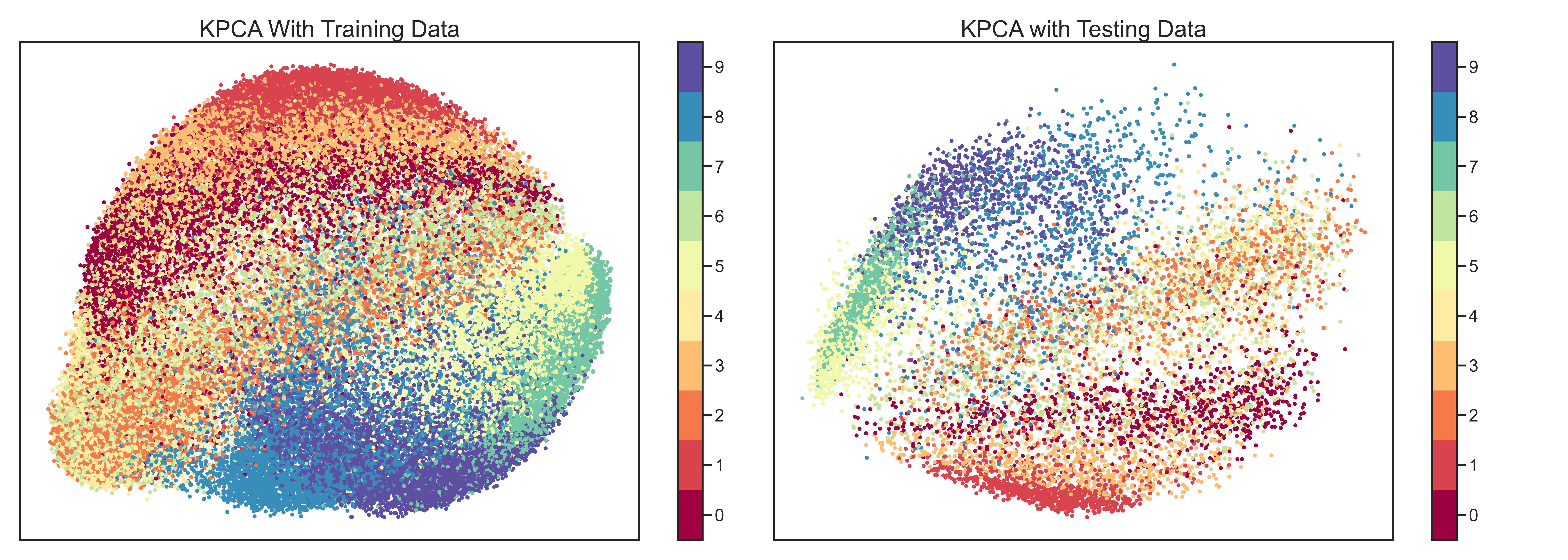}
    \caption{Performance plots of PCA and KPCA}
    \label{fig:PCA&KPCA} 
\end{figure}

\begin{figure}[h!]
    \centering
    \includegraphics[width=0.8\textwidth]{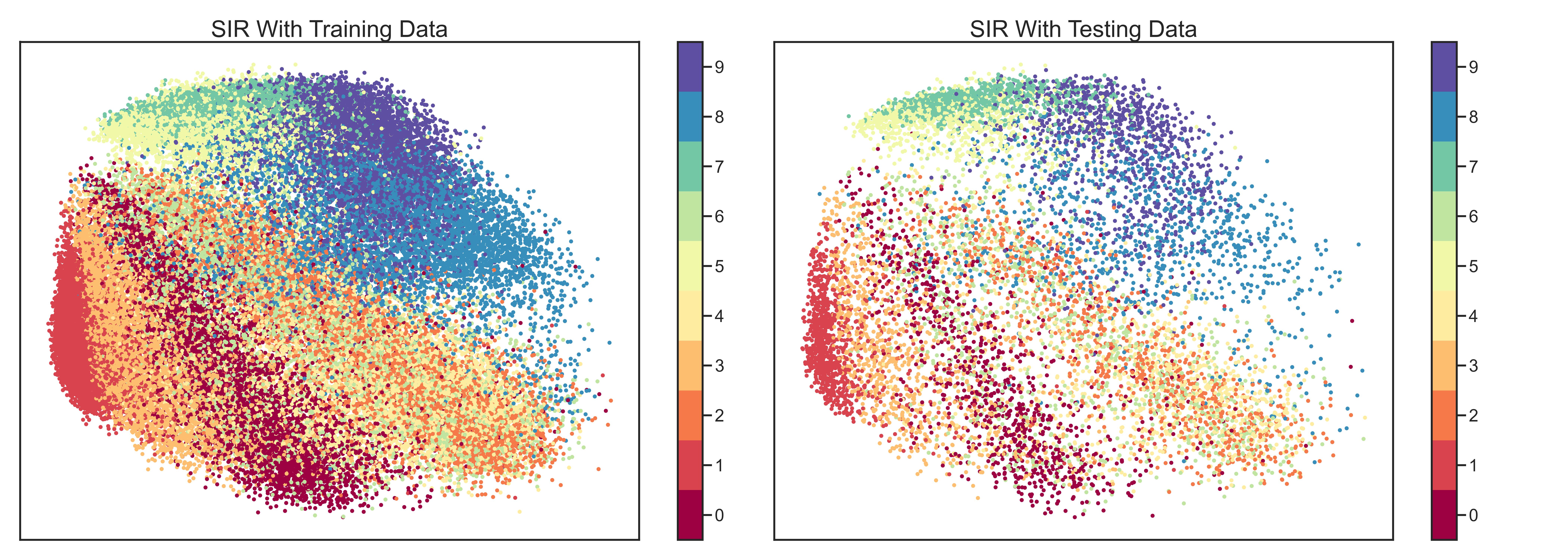}
    \includegraphics[width=0.8\textwidth]{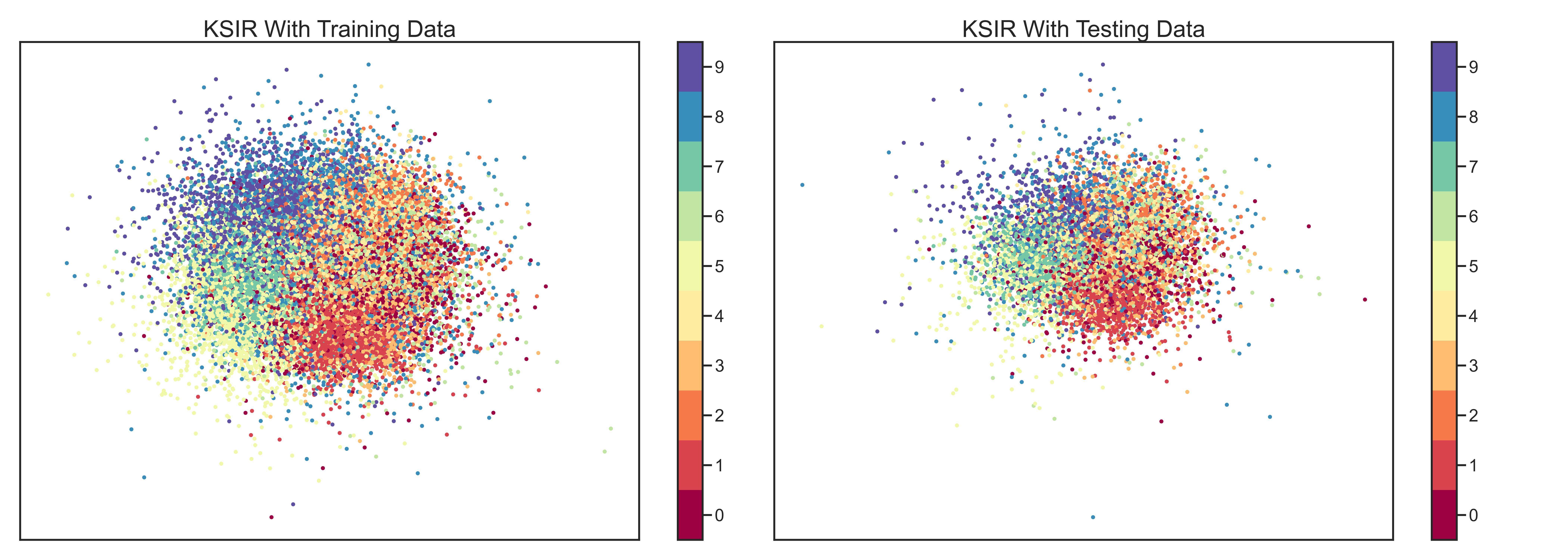}
    \caption{Performance plots of SIR and KSIR}
    \label{fig:SIR&KSIR} 
\end{figure}
t-SNE takes the longest time to run while achieving a strong training score; However, its test score is relatively weak, likely due to a lack of a global transformation that can be adapted to the test dataset.

\begin{figure}[h!]
    \centering
    \includegraphics[width=0.8\textwidth]{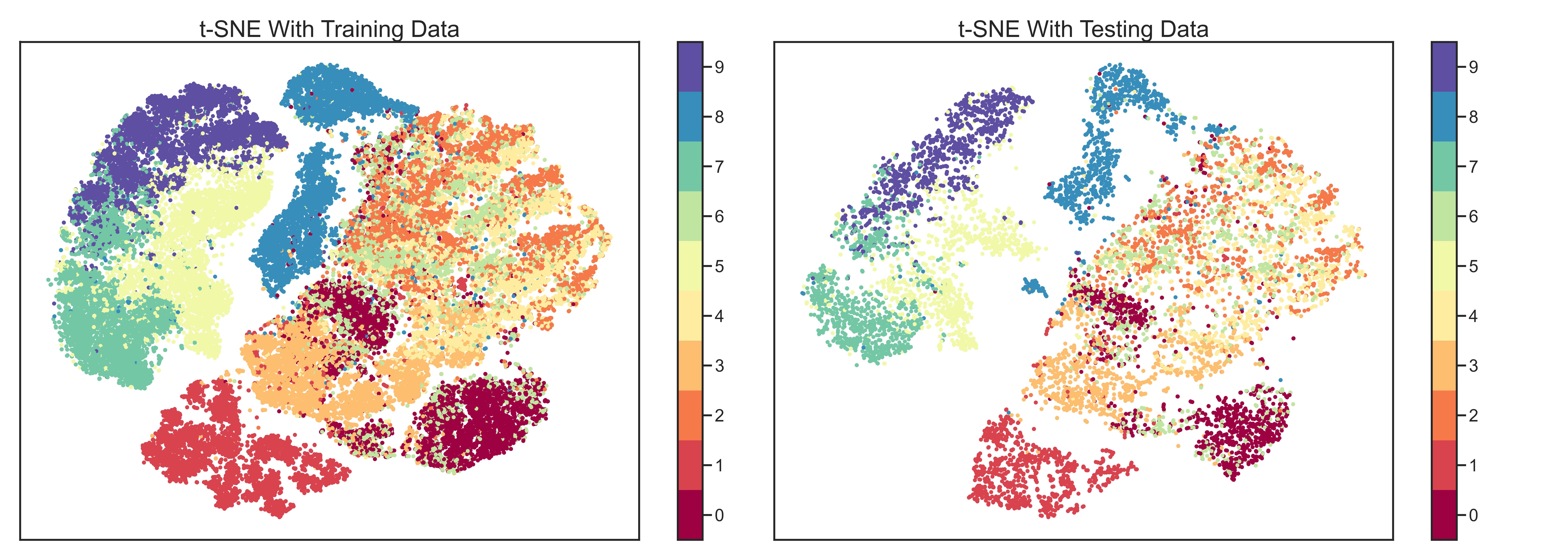}
    \caption{Performance plot of t-SNE}
    \label{fig:TSNE} 
\end{figure}

\subsection{Results for real data with continuous responses}

The second real data studied in our application is the Online News Popularity dataset described in Section 3.2. Table 6 presents the performance comparison on the Online News Popularity dataset. The numerical results are largely consistent with the patterns observed in our simulation studies. Among all the dimension reduction methods, SIR and KSIR achieve the lowest testing MSEs and SEs, and therefore perform the best overall. KSIR, in particular, attains the lowest testing MSE on this dataset. SSU, CaSU, and UU offer moderate improvements over PCA, while CoSU again produces a higher testing MSE than UU and SSU on this numerical response dataset.

The results again indicate the limitations of existing supervised UMAP approaches, which do not fully leverage response information for prediction tasks in regression settings, and motivate further investigation on more effective strategies for integrating response information into nonlinear manifold learning methods for regression.

\begin{table}[h!]
    \centering
    \caption{Results for the Online  News Popularity dataset}
    \label{tab:results_online_news}
    \resizebox{\textwidth}{!}{
        \begin{tabular}{lccccccccccc}
            \toprule
            Type     & Orig   & SSU    & CaSU   & CoSU   & UU     & PCA    & KPCA   & SIR    & KSIR   & t-SNE  \\
            \midrule
            TrainMSE & 0.5097 & 0.4874 & 0.5299 & 0.3860 & 0.5291 & 0.5537 & 0.5771 & 0.5009 & 0.4544 & 0.5212 \\
            TestMSE  & 1.0231 & 1.0946 & 1.0694 & 1.2964 & 1.0702 & 1.1071 & 1.1938 & 1.0049 & 0.9774 & 1.1914 \\
            TestSE   & 0.0009 & 0.0010 & 0.0011 & 0.0013 & 0.0011 & 0.0010 & 0.0397 & 0.0010 & 0.0009 & 0.0027 \\
            \bottomrule
        \end{tabular}
    }
\end{table}

\section{Conclusion}

This study investigates UMAP and other dimension reduction methods, including PCA, SIR, Kernel PCA, Kernel SIR, and t-SNE, in both supervised and unsupervised frameworks for regression and classification tasks and evaluates their performances using simulated and real-world datasets.   KNN was applied to evaluate the prediction performance for both training and testing datasets, with the number of neighbors tuned and random seeds to ensure reproducible results across runs. UMAP exhibited strong performance for classification tasks on simulated datasets and real examples by effectively preserving both local and global data structures, making it a highly effective method for complex datasets with categorical responses. In the simulated settings with continuous responses, supervised linear dimension reduction method SIR achieved better results than the UMAP methods. The supervised UMAP  method CoSU yielded nearly the highest testing MSEs and performed less effectively than the unsupervised UMAP and other dimension methods, suggesting that the current supervised UMAP method did not fully utilize the information from continuous responses during the embedding process. Future work on improving supervised UMAP under the continuous response setting is worthy of further investigation.

Overall, UMAP remains a powerful and versatile tool for dimensionality reduction and offers a strong balance of accuracy and computational efficiency in classification contexts. Its supervised extensions, however, demonstrate limitations in incorporating numerical response information effectively, which points to an important direction for future research on improving supervised UMAP in regression settings.

\section*{Supplementary Information}
Online Resource 1 (ESM\_1.pdf): Additional simulation results for
continuous responses, including KNN regression with three- and
five-dimensional embeddings, ridge regression and random forest with
two-dimensional embeddings, and KNN regression in the
high-dimensional, low-sample-size setting.
\bibliographystyle{asa}
\bibliography{references}
\end{document}


\thispagestyle{empty}

\begin{center}
    {\Large\bfseries Supplementary Information (Online Resource 1)\par}
    \vspace{1em}
    {\large\bfseries A Comparative Study of UMAP and Other Dimensionality Reduction Methods\par}
    \vspace{1em}
    {\itshape Statistics and Computing\par}
    \vspace{2em}
    Guanzhe Zhang$^{1}$, Shanshan Ding$^{1}$, and Zhezhen Jin$^{2}$\par
    \vspace{1em}
    $^{1}$Department of Applied Economics and Statistics, University of Delaware\par
    $^{2}$Department of Biostatistics, Columbia University\par
\end{center}

This document provides additional simulation results for datasets with continuous responses. It includes results for KNN regression using 3- and 5-dimensional embeddings, complementing the 2-dimensional results in the main article, and results obtained using ridge regression and random forest as alternative predictive models. It also reports KNN regression results for 2-dimensional embeddings of high-dimensional, low-sample-size datasets.

\section*{KNN regression with 3-dimensional embeddings}

\begin{table}[!htbp]
\centering
\caption{KNN regression with 3-dimensional embeddings: training and test MSE and standard errors for Model 1 under datasets a1, b1, and c1.}
\label{tab:knn_3d_model1}
\resizebox{\textwidth}{!}{%
\begin{tabular}{llccccccccccc}
\toprule
Model & Type & Orig & SSU & CaSU & CoSU & UU & PCA & KPCA & SIR & KSIR & SPCA & t-SNE \\
\midrule
\multirow{4}{*}{a1} & TrainMSE & 3.3707 & 2.1289 & 3.7361 & 0.2844 & 3.7663 & 3.8171 & 3.8462 & 1.1454 & 0.1053 & 2.4562 & 3.6471 \\
 & TrainSE & 0.0228 & 0.0233 & 0.0291 & 0.0930 & 0.0263 & 0.0270 & 0.0275 & 0.0152 & 0.0019 & 0.0282 & 0.0268 \\
 & TestMSE & 6.8700 & 8.3898 & 7.7111 & 9.6557 & 7.7719 & 7.7478 & 7.7747 & 5.7133 & 11.5926 & 6.1320 & 7.8956 \\
 & TestSE & 0.0374 & 0.0518 & 0.0458 & 0.1365 & 0.0512 & 0.0517 & 0.0550 & 0.0437 & 0.0642 & 0.0473 & 0.0390 \\
\midrule
\multirow{4}{*}{b1} & TrainMSE & 1.8429 & 0.9210 & 1.9880 & 0.2803 & 2.0063 & 2.0635 & 2.0799 & 0.8342 & 0.3006 & 1.4114 & 2.1199 \\
 & TrainSE & 0.0175 & 0.0101 & 0.0180 & 0.0637 & 0.0167 & 0.0174 & 0.0187 & 0.0128 & 0.0079 & 0.0179 & 0.0218 \\
 & TestMSE & 3.6899 & 4.6903 & 4.1356 & 10.8024 & 4.1205 & 4.0500 & 4.1341 & 3.1327 & 54.1122 & 3.2817 & 50.1725 \\
 & TestSE & 0.0299 & 0.0398 & 0.0329 & 0.6498 & 0.0296 & 0.0322 & 0.0351 & 0.0277 & 0.2191 & 0.0371 & 2.4500 \\
\midrule
\multirow{4}{*}{c1} & TrainMSE & 1.1872 & 0.8199 & 1.2748 & 0.2052 & 1.2803 & 1.2411 & 1.2568 & 1.0387 & 0.3070 & 0.8745 & 1.2202 \\
 & TrainSE & 0.0105 & 0.0083 & 0.0099 & 0.0417 & 0.0099 & 0.0111 & 0.0106 & 0.0112 & 0.0074 & 0.0097 & 0.0106 \\
 & TestMSE & 2.3910 & 2.8846 & 2.6390 & 3.9112 & 2.6571 & 2.4986 & 2.5451 & 2.6269 & 17.5752 & 2.0740 & 12.7905 \\
 & TestSE & 0.0184 & 0.0243 & 0.0195 & 0.1510 & 0.0192 & 0.0169 & 0.0185 & 0.0215 & 0.1102 & 0.0191 & 0.7961 \\
\bottomrule
\end{tabular}
}
\end{table}

\begin{table}[!htbp]
\centering
\caption{KNN regression with 3-dimensional embeddings: training and test MSE and standard errors for Model 2 under datasets a2, b2, and c2.}
\label{tab:knn_3d_model2}
\resizebox{\textwidth}{!}{%
\begin{tabular}{llccccccccccc}
\toprule
Model & Type & Orig & SSU & CaSU & CoSU & UU & PCA & KPCA & SIR & KSIR & SPCA & t-SNE \\
\midrule
\multirow{4}{*}{a2} & TrainMSE & 1.8105 & 1.0994 & 1.9973 & 0.1980 & 2.0009 & 2.0916 & 2.0965 & 0.5251 & 0.1018 & 1.1581 & 1.9306 \\
 & TrainSE & 0.0166 & 0.0131 & 0.0177 & 0.0567 & 0.0186 & 0.0188 & 0.0210 & 0.0060 & 0.0027 & 0.0115 & 0.0183 \\
 & TestMSE & 3.6614 & 4.5473 & 4.2162 & 5.1759 & 4.2128 & 4.2342 & 4.2807 & 2.5736 & 6.2790 & 3.0328 & 4.3132 \\
 & TestSE & 0.0257 & 0.0330 & 0.0325 & 0.0786 & 0.0301 & 0.0334 & 0.0454 & 0.0202 & 0.0467 & 0.0272 & 0.0336 \\
\midrule
\multirow{4}{*}{b2} & TrainMSE & 2.5255 & 1.2752 & 2.8251 & 0.1048 & 2.7557 & 2.8752 & 2.9087 & 0.9679 & 0.3653 & 1.6420 & 2.8298 \\
 & TrainSE & 0.0237 & 0.0169 & 0.0297 & 0.0402 & 0.0267 & 0.0314 & 0.0314 & 0.0160 & 0.0080 & 0.0180 & 0.0292 \\
 & TestMSE & 5.1633 & 6.5278 & 5.9658 & 13.3977 & 5.9347 & 5.7681 & 5.9247 & 4.1328 & 58.3585 & 4.1444 & 49.2849 \\
 & TestSE & 0.0469 & 0.0526 & 0.0500 & 0.7181 & 0.0524 & 0.0561 & 0.0606 & 0.0479 & 0.2962 & 0.0526 & 2.4695 \\
\midrule
\multirow{4}{*}{c2} & TrainMSE & 1.4017 & 0.9745 & 1.4985 & 0.1855 & 1.5208 & 1.4539 & 1.4737 & 1.1884 & 0.5740 & 0.9375 & 1.4247 \\
 & TrainSE & 0.0134 & 0.0120 & 0.0132 & 0.0418 & 0.0142 & 0.0138 & 0.0154 & 0.0157 & 0.0174 & 0.0141 & 0.0155 \\
 & TestMSE & 2.8682 & 3.4512 & 3.2092 & 5.0333 & 3.2173 & 3.0112 & 3.0496 & 3.1819 & 26.1652 & 2.3397 & 18.6563 \\
 & TestSE & 0.0265 & 0.0306 & 0.0300 & 0.2339 & 0.0297 & 0.0284 & 0.0267 & 0.0285 & 0.1812 & 0.0312 & 1.1428 \\
\bottomrule
\end{tabular}
}
\end{table}

\begin{table}[!htbp]
\centering
\caption{KNN regression with 3-dimensional embeddings: training and test MSE and standard errors for Model 3 under datasets a3, b3, and c3.}
\label{tab:knn_3d_model3}
\resizebox{\textwidth}{!}{%
\begin{tabular}{llccccccccccc}
\toprule
Model & Type & Orig & SSU & CaSU & CoSU & UU & PCA & KPCA & SIR & KSIR & SPCA & t-SNE \\
\midrule
\multirow{4}{*}{a3} & TrainMSE & 0.4142 & 0.2436 & 0.4334 & 0.0246 & 0.4344 & 0.4486 & 0.4472 & 0.1395 & 0.0125 & 0.2949 & 0.4293 \\
 & TrainSE & 0.0032 & 0.0027 & 0.0038 & 0.0089 & 0.0036 & 0.0036 & 0.0036 & 0.0018 & 0.0002 & 0.0027 & 0.0035 \\
 & TestMSE & 0.8314 & 1.0072 & 0.9144 & 1.2024 & 0.9099 & 0.9112 & 0.9189 & 0.6945 & 1.3494 & 0.7542 & 0.9248 \\
 & TestSE & 0.0053 & 0.0073 & 0.0058 & 0.0212 & 0.0062 & 0.0065 & 0.0079 & 0.0056 & 0.0090 & 0.0059 & 0.0059 \\
\midrule
\multirow{4}{*}{b3} & TrainMSE & 0.4174 & 0.2367 & 0.4335 & 0.0166 & 0.4342 & 0.4463 & 0.4489 & 0.1728 & 0.0128 & 0.2847 & 0.4258 \\
 & TrainSE & 0.0037 & 0.0025 & 0.0036 & 0.0074 & 0.0037 & 0.0035 & 0.0036 & 0.0026 & 0.0002 & 0.0056 & 0.0034 \\
 & TestMSE & 0.8397 & 0.9991 & 0.9091 & 1.2238 & 0.9065 & 0.8970 & 0.9156 & 0.7044 & 1.3618 & 0.6591 & 0.9251 \\
 & TestSE & 0.0048 & 0.0063 & 0.0058 & 0.0215 & 0.0065 & 0.0057 & 0.0073 & 0.0061 & 0.0079 & 0.0103 & 0.0053 \\
\midrule
\multirow{4}{*}{c3} & TrainMSE & 0.3076 & 0.1912 & 0.3172 & 0.0393 & 0.3198 & 0.3192 & 0.3191 & 0.2779 & 0.0126 & 0.3128 & 0.3114 \\
 & TrainSE & 0.0026 & 0.0022 & 0.0026 & 0.0099 & 0.0026 & 0.0027 & 0.0026 & 0.0036 & 0.0002 & 0.0026 & 0.0025 \\
 & TestMSE & 0.6215 & 0.7253 & 0.6519 & 0.8752 & 0.6514 & 0.6354 & 0.6358 & 0.6521 & 1.4921 & 0.6321 & 1.0955 \\
 & TestSE & 0.0041 & 0.0061 & 0.0045 & 0.0182 & 0.0049 & 0.0050 & 0.0045 & 0.0051 & 0.0097 & 0.0045 & 0.0277 \\
\bottomrule
\end{tabular}
}
\end{table}\FloatBarrier
\section*{KNN regression with 5-dimensional embeddings}

\begin{table}[!htbp]
\centering
\caption{KNN regression with 5-dimensional embeddings: training and test MSE and standard errors for Model 1 under datasets a1, b1, and c1.}
\label{tab:knn_5d_model1}
\resizebox{\textwidth}{!}{%
\begin{tabular}{llccccccccccc}
\toprule
Model & Type & Orig & SSU & CaSU & CoSU & UU & PCA & KPCA & SIR & KSIR & SPCA & t-SNE \\
\midrule
\multirow{4}{*}{a1} & TrainMSE & 3.3707 & 2.1208 & 3.6591 & 0.2620 & 3.6645 & 3.8006 & 3.8582 & 0.8955 & 0.1035 & 2.3890 & 3.5736 \\
 & TrainSE & 0.0228 & 0.0224 & 0.0268 & 0.0912 & 0.0252 & 0.0259 & 0.0279 & 0.0127 & 0.0019 & 0.0200 & 0.0252 \\
 & TestMSE & 6.8700 & 8.3919 & 7.6520 & 9.3356 & 7.6612 & 7.7179 & 7.7604 & 5.9229 & 11.5976 & 6.0484 & 7.8577 \\
 & TestSE & 0.0374 & 0.0465 & 0.0444 & 0.1092 & 0.0478 & 0.0494 & 0.0594 & 0.0455 & 0.0637 & 0.0473 & 0.0467 \\
\midrule
\multirow{4}{*}{b1} & TrainMSE & 1.8429 & 0.9213 & 1.9551 & 0.2574 & 1.9808 & 2.0462 & 2.0725 & 0.6300 & 0.2976 & 1.4115 & 2.3569 \\
 & TrainSE & 0.0175 & 0.0109 & 0.0165 & 0.0616 & 0.0152 & 0.0169 & 0.0171 & 0.0105 & 0.0077 & 0.0165 & 0.0292 \\
 & TestMSE & 3.6899 & 4.6777 & 4.0774 & 8.2423 & 4.0744 & 4.0928 & 4.1158 & 3.3357 & 54.0910 & 3.2622 & 46.4636 \\
 & TestSE & 0.0299 & 0.0398 & 0.0320 & 0.4771 & 0.0320 & 0.0356 & 0.0441 & 0.0304 & 0.2204 & 0.0379 & 1.7434 \\
\midrule
\multirow{4}{*}{c1} & TrainMSE & 1.1872 & 0.8237 & 1.2655 & 0.1988 & 1.2716 & 1.2407 & 1.2557 & 0.8613 & 0.3081 & 0.8842 & 1.2250 \\
 & TrainSE & 0.0105 & 0.0085 & 0.0102 & 0.0406 & 0.0116 & 0.0102 & 0.0102 & 0.0102 & 0.0078 & 0.0101 & 0.0120 \\
 & TestMSE & 2.3910 & 2.8789 & 2.6205 & 3.6041 & 2.6194 & 2.5218 & 2.5363 & 2.6754 & 17.5593 & 2.0916 & 11.6787 \\
 & TestSE & 0.0184 & 0.0245 & 0.0193 & 0.1125 & 0.0184 & 0.0201 & 0.0187 & 0.0223 & 0.1082 & 0.0200 & 0.5627 \\
\bottomrule
\end{tabular}
}
\end{table}

\begin{table}[!htbp]
\centering
\caption{KNN regression with 5-dimensional embeddings: training and test MSE and standard errors for Model 2 under datasets a2, b2, and c2.}
\label{tab:knn_5d_model2}
\resizebox{\textwidth}{!}{%
\begin{tabular}{llccccccccccc}
\toprule
Model & Type & Orig & SSU & CaSU & CoSU & UU & PCA & KPCA & SIR & KSIR & SPCA & t-SNE \\
\midrule
\multirow{4}{*}{a2} & TrainMSE & 1.8105 & 1.0856 & 1.9574 & 0.1853 & 1.9706 & 2.0752 & 2.0823 & 0.4156 & 0.1024 & 1.1610 & 1.9043 \\
 & TrainSE & 0.0166 & 0.0123 & 0.0186 & 0.0553 & 0.0177 & 0.0194 & 0.0197 & 0.0050 & 0.0028 & 0.0113 & 0.0174 \\
 & TestMSE & 3.6614 & 4.5456 & 4.1974 & 5.1787 & 4.1766 & 4.1974 & 4.2233 & 2.7221 & 6.2812 & 2.9789 & 4.3092 \\
 & TestSE & 0.0257 & 0.0325 & 0.0276 & 0.0839 & 0.0317 & 0.0373 & 0.0526 & 0.0224 & 0.0475 & 0.0270 & 0.0318 \\
\midrule
\multirow{4}{*}{b2} & TrainMSE & 2.5255 & 1.2631 & 2.7255 & 0.0773 & 2.7415 & 2.8143 & 2.8751 & 0.7408 & 0.3716 & 1.7180 & 3.1422 \\
 & TrainSE & 0.0237 & 0.0178 & 0.0268 & 0.0327 & 0.0275 & 0.0293 & 0.0307 & 0.0154 & 0.0091 & 0.0219 & 0.0311 \\
 & TestMSE & 5.1633 & 6.5739 & 5.8223 & 11.1983 & 5.8447 & 5.7197 & 5.9619 & 4.2732 & 58.3596 & 4.0790 & 50.3167 \\
 & TestSE & 0.0469 & 0.0583 & 0.0467 & 0.5720 & 0.0439 & 0.0452 & 0.0765 & 0.0431 & 0.2985 & 0.0549 & 1.6631 \\
\midrule
\multirow{4}{*}{c2} & TrainMSE & 1.4017 & 0.9745 & 1.4965 & 0.1811 & 1.5082 & 1.4607 & 1.4626 & 0.9824 & 0.5846 & 0.9405 & 1.4671 \\
 & TrainSE & 0.0134 & 0.0115 & 0.0142 & 0.0405 & 0.0143 & 0.0151 & 0.0137 & 0.0126 & 0.0178 & 0.0145 & 0.0174 \\
 & TestMSE & 2.8682 & 3.4777 & 3.1790 & 4.7387 & 3.1919 & 3.0320 & 3.0603 & 3.2243 & 26.1641 & 2.3702 & 17.3651 \\
 & TestSE & 0.0265 & 0.0357 & 0.0311 & 0.1924 & 0.0266 & 0.0285 & 0.0258 & 0.0296 & 0.1800 & 0.0341 & 0.8122 \\
\bottomrule
\end{tabular}
}
\end{table}

\begin{table}[!htbp]
\centering
\caption{KNN regression with 5-dimensional embeddings: training and test MSE and standard errors for Model 3 under datasets a3, b3, and c3.}
\label{tab:knn_5d_model3}
\resizebox{\textwidth}{!}{%
\begin{tabular}{llccccccccccc}
\toprule
Model & Type & Orig & SSU & CaSU & CoSU & UU & PCA & KPCA & SIR & KSIR & SPCA & t-SNE \\
\midrule
\multirow{4}{*}{a3} & TrainMSE & 0.4142 & 0.2433 & 0.4346 & 0.0216 & 0.4313 & 0.4469 & 0.4452 & 0.1077 & 0.0124 & 0.2957 & 0.4223 \\
 & TrainSE & 0.0032 & 0.0029 & 0.0034 & 0.0086 & 0.0033 & 0.0035 & 0.0038 & 0.0016 & 0.0002 & 0.0025 & 0.0037 \\
 & TestMSE & 0.8314 & 0.9956 & 0.9079 & 1.1701 & 0.9098 & 0.9142 & 0.9079 & 0.7131 & 1.3489 & 0.7453 & 0.9261 \\
 & TestSE & 0.0053 & 0.0066 & 0.0057 & 0.0166 & 0.0061 & 0.0071 & 0.0094 & 0.0055 & 0.0089 & 0.0063 & 0.0058 \\
\midrule
\multirow{4}{*}{b3} & TrainMSE & 0.4174 & 0.2350 & 0.4328 & 0.0147 & 0.4341 & 0.4413 & 0.4455 & 0.1403 & 0.0128 & 0.2799 & 0.4238 \\
 & TrainSE & 0.0037 & 0.0024 & 0.0036 & 0.0072 & 0.0037 & 0.0038 & 0.0039 & 0.0025 & 0.0002 & 0.0043 & 0.0036 \\
 & TestMSE & 0.8397 & 0.9904 & 0.9005 & 1.2122 & 0.8991 & 0.8992 & 0.9065 & 0.7209 & 1.3616 & 0.6535 & 0.9185 \\
 & TestSE & 0.0048 & 0.0062 & 0.0057 & 0.0208 & 0.0055 & 0.0064 & 0.0079 & 0.0060 & 0.0079 & 0.0093 & 0.0060 \\
\midrule
\multirow{4}{*}{c3} & TrainMSE & 0.3076 & 0.1890 & 0.3162 & 0.0381 & 0.3163 & 0.3146 & 0.3192 & 0.2374 & 0.0126 & 0.3096 & 0.3091 \\
 & TrainSE & 0.0026 & 0.0021 & 0.0025 & 0.0097 & 0.0026 & 0.0026 & 0.0026 & 0.0031 & 0.0002 & 0.0027 & 0.0026 \\
 & TestMSE & 0.6215 & 0.7198 & 0.6456 & 0.8452 & 0.6483 & 0.6352 & 0.6362 & 0.6640 & 1.4947 & 0.6321 & 1.1133 \\
 & TestSE & 0.0041 & 0.0056 & 0.0046 & 0.0151 & 0.0045 & 0.0047 & 0.0050 & 0.0048 & 0.0097 & 0.0045 & 0.0217 \\
\bottomrule
\end{tabular}
}
\end{table}\FloatBarrier
\section*{Ridge regression with 2-dimensional embeddings}

\begin{table}[!htbp]
\centering
\caption{Ridge regression with 2-dimensional embeddings: training and test MSE and standard errors for Model 1 under datasets a1, b1, and c1.}
\label{tab:ridge_2d_model1}
\resizebox{\textwidth}{!}{%
\begin{tabular}{llccccccccccc}
\toprule
Model & Type & Orig & SSU & CaSU & CoSU & UU & PCA & KPCA & SIR & KSIR & SPCA & t-SNE \\
\midrule
\multirow{4}{*}{a1} & TrainMSE & 1.3016 & 5.5828 & 5.8051 & 4.8040 & 5.8079 & 5.7694 & 5.7752 & 2.1227 & 4.6638 & 3.7200 & 5.8034 \\
 & TrainSE & 0.0297 & 0.0330 & 0.0311 & 0.1061 & 0.0308 & 0.0305 & 0.0304 & 0.0197 & 0.0873 & 0.0386 & 0.0309 \\
 & TestMSE & 4.3311 & 5.9731 & 5.8859 & 6.0833 & 5.8794 & 5.8568 & 5.8599 & 4.6709 & 7.0124 & 4.9676 & 5.8765 \\
 & TestSE & 0.0295 & 0.0297 & 0.0253 & 0.0343 & 0.0245 & 0.0239 & 0.0240 & 0.0276 & 0.0901 & 0.0355 & 0.0244 \\
\midrule
\multirow{4}{*}{b1} & TrainMSE & 0.5776 & 3.9272 & 3.3120 & 15.2969 & 3.4172 & 3.0661 & 3.2630 & 1.4781 & 21.3609 & 2.1439 & 3.3908 \\
 & TrainSE & 0.0225 & 0.1086 & 0.0237 & 0.9715 & 0.0245 & 0.0207 & 0.0220 & 0.0223 & 0.4123 & 0.0245 & 0.0237 \\
 & TestMSE & 2.1876 & 3.8131 & 3.6372 & 14.5380 & 3.6690 & 3.0622 & 3.2851 & 2.5069 & 32.9108 & 2.6607 & 45.5942 \\
 & TestSE & 0.0189 & 0.0389 & 0.0321 & 0.9660 & 0.0279 & 0.0244 & 0.0253 & 0.0205 & 0.4276 & 0.0304 & 3.6242 \\
\midrule
\multirow{4}{*}{c1} & TrainMSE & 0.5567 & 2.3209 & 2.1947 & 4.5589 & 2.2266 & 1.8704 & 3.1748 & 1.8552 & 7.1419 & 1.2902 & 2.1850 \\
 & TrainSE & 0.0076 & 0.0585 & 0.0162 & 0.2750 & 0.0157 & 0.0125 & 0.0315 & 0.0122 & 0.1602 & 0.0102 & 0.0160 \\
 & TestMSE & 1.4804 & 2.7023 & 2.2658 & 5.0238 & 2.2964 & 1.9034 & 3.2612 & 1.8984 & 10.8042 & 1.5959 & 13.6080 \\
 & TestSE & 0.0110 & 0.0475 & 0.0184 & 0.2339 & 0.0192 & 0.0138 & 0.0354 & 0.0137 & 0.1749 & 0.0151 & 1.0198 \\
\bottomrule
\end{tabular}
}
\end{table}

\begin{table}[!htbp]
\centering
\caption{Ridge regression with 2-dimensional embeddings: training and test MSE and standard errors for Model 2 under datasets a2, b2, and c2.}
\label{tab:ridge_2d_model2}
\resizebox{\textwidth}{!}{%
\begin{tabular}{llccccccccccc}
\toprule
Model & Type & Orig & SSU & CaSU & CoSU & UU & PCA & KPCA & SIR & KSIR & SPCA & t-SNE \\
\midrule
\multirow{4}{*}{a2} & TrainMSE & 0.2681 & 2.9998 & 3.1702 & 2.6385 & 3.1693 & 3.1455 & 3.1454 & 0.9736 & 2.5029 & 1.7916 & 3.1657 \\
 & TrainSE & 0.0122 & 0.0252 & 0.0231 & 0.0529 & 0.0231 & 0.0232 & 0.0228 & 0.0075 & 0.0564 & 0.0158 & 0.0233 \\
 & TestMSE & 1.7468 & 3.2596 & 3.2044 & 3.3071 & 3.1962 & 3.1812 & 3.1826 & 2.1690 & 3.8427 & 2.4607 & 3.1966 \\
 & TestSE & 0.0134 & 0.0238 & 0.0240 & 0.0277 & 0.0236 & 0.0234 & 0.0234 & 0.0160 & 0.0590 & 0.0216 & 0.0232 \\
\midrule
\multirow{4}{*}{b2} & TrainMSE & 1.2707 & 4.6697 & 4.6266 & 17.2749 & 4.6924 & 4.3395 & 4.4810 & 2.1661 & 23.2245 & 3.0250 & 4.6874 \\
 & TrainSE & 0.0347 & 0.1287 & 0.0380 & 0.9888 & 0.0385 & 0.0363 & 0.0354 & 0.0290 & 0.4711 & 0.0219 & 0.0396 \\
 & TestMSE & 3.4476 & 5.1463 & 5.0549 & 16.4664 & 5.0840 & 4.4356 & 4.5884 & 3.8101 & 35.8795 & 3.7273 & 50.3267 \\
 & TestSE & 0.0271 & 0.0463 & 0.0390 & 0.9791 & 0.0457 & 0.0353 & 0.0372 & 0.0332 & 0.4919 & 0.0353 & 3.8696 \\
\midrule
\multirow{4}{*}{c2} & TrainMSE & 1.2757 & 3.1936 & 3.3134 & 7.2165 & 3.1812 & 3.3933 & 4.0761 & 3.3178 & 10.1808 & 2.4884 & 3.7322 \\
 & TrainSE & 0.0339 & 0.0633 & 0.0433 & 0.4000 & 0.0404 & 0.0297 & 0.0548 & 0.0301 & 0.2781 & 0.0249 & 0.0409 \\
 & TestMSE & 3.0423 & 3.9509 & 3.7178 & 7.6144 & 3.5102 & 3.5262 & 4.3174 & 3.5551 & 16.5331 & 3.1206 & 19.5989 \\
 & TestSE & 0.0307 & 0.0625 & 0.0487 & 0.3400 & 0.0490 & 0.0327 & 0.0682 & 0.0339 & 0.2851 & 0.0365 & 1.5562 \\
\bottomrule
\end{tabular}
}
\end{table}

\begin{table}[!htbp]
\centering
\caption{Ridge regression with 2-dimensional embeddings: training and test MSE and standard errors for Model 3 under datasets a3, b3, and c3.}
\label{tab:ridge_2d_model3}
\resizebox{\textwidth}{!}{%
\begin{tabular}{llccccccccccc}
\toprule
Model & Type & Orig & SSU & CaSU & CoSU & UU & PCA & KPCA & SIR & KSIR & SPCA & t-SNE \\
\midrule
\multirow{4}{*}{a3} & TrainMSE & 0.1879 & 0.6398 & 0.6743 & 0.5384 & 0.6743 & 0.6711 & 0.6712 & 0.2706 & 0.5348 & 0.4529 & 0.6734 \\
 & TrainSE & 0.0062 & 0.0054 & 0.0045 & 0.0140 & 0.0044 & 0.0043 & 0.0044 & 0.0022 & 0.0110 & 0.0050 & 0.0044 \\
 & TestMSE & 0.5414 & 0.7096 & 0.6901 & 0.7211 & 0.6902 & 0.6870 & 0.6872 & 0.5678 & 0.8291 & 0.6023 & 0.6902 \\
 & TestSE & 0.0040 & 0.0042 & 0.0036 & 0.0051 & 0.0037 & 0.0036 & 0.0036 & 0.0041 & 0.0118 & 0.0050 & 0.0036 \\
\midrule
\multirow{4}{*}{b3} & TrainMSE & 0.2141 & 0.6629 & 0.6751 & 0.6056 & 0.6748 & 0.6731 & 0.6734 & 0.3034 & 0.5120 & 0.5882 & 0.6747 \\
 & TrainSE & 0.0051 & 0.0040 & 0.0041 & 0.0095 & 0.0041 & 0.0041 & 0.0041 & 0.0049 & 0.0131 & 0.0092 & 0.0041 \\
 & TestMSE & 0.5511 & 0.6928 & 0.6888 & 0.7034 & 0.6886 & 0.6862 & 0.6863 & 0.5821 & 0.8538 & 0.6499 & 0.7140 \\
 & TestSE & 0.0035 & 0.0039 & 0.0038 & 0.0043 & 0.0038 & 0.0039 & 0.0038 & 0.0041 & 0.0141 & 0.0064 & 0.0111 \\
\midrule
\multirow{4}{*}{c3} & TrainMSE & 0.6420 & 0.6324 & 0.6571 & 0.3994 & 0.6570 & 0.7031 & 0.6066 & 0.6878 & 0.5716 & 0.6855 & 0.6666 \\
 & TrainSE & 0.0055 & 0.0045 & 0.0039 & 0.0194 & 0.0038 & 0.0035 & 0.0034 & 0.0040 & 0.0127 & 0.0036 & 0.0036 \\
 & TestMSE & 0.7188 & 0.6661 & 0.6863 & 0.6323 & 0.6810 & 0.7169 & 0.6247 & 0.7226 & 0.9302 & 0.7204 & 0.8722 \\
 & TestSE & 0.0040 & 0.0043 & 0.0042 & 0.0068 & 0.0044 & 0.0041 & 0.0042 & 0.0043 & 0.0136 & 0.0043 & 0.0244 \\
\bottomrule
\end{tabular}
}
\end{table}

\FloatBarrier

\section*{Random forest with 2-dimensional embeddings}

\begin{table}[!htbp]
\centering
\caption{Random forest with 2-dimensional embeddings: training and test MSE and standard errors for Model 1 under datasets a1, b1, and c1.}
\label{tab:rf_2d_model1}
\resizebox{\textwidth}{!}{%
\begin{tabular}{llccccccccccc}
\toprule
Model & Type & Orig & SSU & CaSU & CoSU & UU & PCA & KPCA & SIR & KSIR & SPCA & t-SNE \\
\midrule
\multirow{4}{*}{a1} & TrainMSE & 1.2925 & 1.2996 & 3.9122 & 0.3161 & 3.8743 & 3.9270 & 3.9021 & 1.2800 & 0.1535 & 2.3407 & 3.8406 \\
 & TrainSE & 0.0458 & 0.0526 & 0.0327 & 0.1028 & 0.0363 & 0.0389 & 0.0359 & 0.0206 & 0.0025 & 0.0249 & 0.0438 \\
 & TestMSE & 5.1859 & 7.0436 & 6.3530 & 8.1604 & 6.3329 & 6.1932 & 6.2257 & 4.9912 & 11.5382 & 5.4117 & 6.3318 \\
 & TestSE & 0.0230 & 0.0458 & 0.0455 & 0.0979 & 0.0355 & 0.0305 & 0.0305 & 0.0345 & 0.0628 & 0.0383 & 0.0378 \\
\midrule
\multirow{4}{*}{b1} & TrainMSE & 0.5684 & 0.6759 & 2.1084 & 0.2902 & 2.0851 & 2.0864 & 2.1483 & 0.9826 & 0.4403 & 1.3450 & 2.0358 \\
 & TrainSE & 0.0179 & 0.0360 & 0.0219 & 0.0630 & 0.0251 & 0.0280 & 0.0195 & 0.0237 & 0.0108 & 0.0333 & 0.0319 \\
 & TestMSE & 3.2573 & 4.0346 & 4.0521 & 10.2983 & 3.7164 & 3.2826 & 3.3444 & 2.6849 & 53.9761 & 2.8252 & 50.8777 \\
 & TestSE & 0.0253 & 0.0376 & 0.0830 & 0.6382 & 0.0447 & 0.0249 & 0.0274 & 0.0233 & 0.2187 & 0.0296 & 3.5371 \\
\midrule
\multirow{4}{*}{c1} & TrainMSE & 0.4118 & 0.6466 & 1.3231 & 0.2118 & 1.3192 & 1.1651 & 1.0653 & 1.0171 & 0.4469 & 0.5875 & 1.2698 \\
 & TrainSE & 0.0186 & 0.0357 & 0.0234 & 0.0457 & 0.0261 & 0.0380 & 0.0490 & 0.0390 & 0.0098 & 0.0299 & 0.0246 \\
 & TestMSE & 2.1336 & 2.6087 & 2.3798 & 3.6792 & 2.3525 & 2.2022 & 2.3935 & 2.3474 & 17.4405 & 1.8753 & 12.2816 \\
 & TestSE & 0.0177 & 0.0239 & 0.0248 & 0.1320 & 0.0248 & 0.0163 & 0.0240 & 0.0205 & 0.1119 & 0.0167 & 0.7969 \\
\bottomrule
\end{tabular}
}
\end{table}

\begin{table}[!htbp]
\centering
\caption{Random forest with 2-dimensional embeddings: training and test MSE and standard errors for Model 2 under datasets a2, b2, and c2.}
\label{tab:rf_2d_model2}
\resizebox{\textwidth}{!}{%
\begin{tabular}{llccccccccccc}
\toprule
Model & Type & Orig & SSU & CaSU & CoSU & UU & PCA & KPCA & SIR & KSIR & SPCA & t-SNE \\
\midrule
\multirow{4}{*}{a2} & TrainMSE & 0.6136 & 0.6415 & 2.1299 & 0.2088 & 2.1299 & 2.1255 & 2.1535 & 0.5851 & 0.1529 & 1.1484 & 2.1381 \\
 & TrainSE & 0.0257 & 0.0364 & 0.0226 & 0.0590 & 0.0196 & 0.0215 & 0.0229 & 0.0103 & 0.0036 & 0.0122 & 0.0204 \\
 & TestMSE & 2.6805 & 3.8530 & 3.4693 & 4.3961 & 3.4515 & 3.3722 & 3.3591 & 2.2772 & 6.2285 & 2.6519 & 3.4352 \\
 & TestSE & 0.0200 & 0.0277 & 0.0278 & 0.0613 & 0.0278 & 0.0264 & 0.0253 & 0.0170 & 0.0461 & 0.0229 & 0.0262 \\
\midrule
\multirow{4}{*}{b2} & TrainMSE & 0.7334 & 0.9507 & 2.8848 & 0.1398 & 2.8918 & 2.9511 & 2.8824 & 1.0127 & 0.5374 & 1.5036 & 2.8872 \\
 & TrainSE & 0.0247 & 0.0524 & 0.0425 & 0.0463 & 0.0449 & 0.0441 & 0.0487 & 0.0413 & 0.0117 & 0.0433 & 0.0333 \\
 & TestMSE & 4.3119 & 5.6961 & 5.6275 & 12.0605 & 5.4705 & 4.7162 & 4.8518 & 3.6875 & 58.1718 & 3.6792 & 50.7176 \\
 & TestSE & 0.0381 & 0.0559 & 0.0743 & 0.6377 & 0.0709 & 0.0399 & 0.0449 & 0.0398 & 0.2918 & 0.0553 & 3.6519 \\
\midrule
\multirow{4}{*}{c2} & TrainMSE & 0.4378 & 0.8037 & 1.4945 & 0.1552 & 1.5596 & 1.2296 & 1.2761 & 1.0852 & 0.8552 & 0.5221 & 1.3685 \\
 & TrainSE & 0.0183 & 0.0447 & 0.0407 & 0.0398 & 0.0385 & 0.0551 & 0.0619 & 0.0517 & 0.0244 & 0.0332 & 0.0451 \\
 & TestMSE & 2.7250 & 3.2391 & 3.0578 & 4.7231 & 2.9605 & 2.7145 & 3.1500 & 3.0061 & 25.8731 & 2.2053 & 16.9073 \\
 & TestSE & 0.0317 & 0.0388 & 0.0458 & 0.1974 & 0.0322 & 0.0296 & 0.0634 & 0.0525 & 0.1811 & 0.0339 & 1.1083 \\
\bottomrule
\end{tabular}
}
\end{table}

\begin{table}[!htbp]
\centering
\caption{Random forest with 2-dimensional embeddings: training and test MSE and standard errors for Model 3 under datasets a3, b3, and c3.}
\label{tab:rf_2d_model3}
\resizebox{\textwidth}{!}{%
\begin{tabular}{llccccccccccc}
\toprule
Model & Type & Orig & SSU & CaSU & CoSU & UU & PCA & KPCA & SIR & KSIR & SPCA & t-SNE \\
\midrule
\multirow{4}{*}{a3} & TrainMSE & 0.1509 & 0.1370 & 0.4508 & 0.0277 & 0.4448 & 0.4533 & 0.4585 & 0.1589 & 0.0186 & 0.2851 & 0.4503 \\
 & TrainSE & 0.0053 & 0.0068 & 0.0051 & 0.0101 & 0.0055 & 0.0053 & 0.0050 & 0.0018 & 0.0003 & 0.0023 & 0.0051 \\
 & TestMSE & 0.6211 & 0.8410 & 0.7450 & 0.9976 & 0.7502 & 0.7245 & 0.7265 & 0.6050 & 1.3433 & 0.6435 & 0.7365 \\
 & TestSE & 0.0034 & 0.0062 & 0.0049 & 0.0128 & 0.0063 & 0.0042 & 0.0041 & 0.0044 & 0.0090 & 0.0050 & 0.0041 \\
\midrule
\multirow{4}{*}{b3} & TrainMSE & 0.1791 & 0.1395 & 0.4516 & 0.0200 & 0.4534 & 0.4443 & 0.4587 & 0.1924 & 0.0187 & 0.2877 & 0.4493 \\
 & TrainSE & 0.0050 & 0.0078 & 0.0052 & 0.0082 & 0.0048 & 0.0058 & 0.0047 & 0.0028 & 0.0003 & 0.0056 & 0.0049 \\
 & TestMSE & 0.6454 & 0.8344 & 0.7417 & 1.0186 & 0.7434 & 0.7186 & 0.7298 & 0.6071 & 1.3567 & 0.6034 & 0.7601 \\
 & TestSE & 0.0037 & 0.0057 & 0.0050 & 0.0120 & 0.0051 & 0.0043 & 0.0042 & 0.0048 & 0.0079 & 0.0117 & 0.0099 \\
\midrule
\multirow{4}{*}{c3} & TrainMSE & 0.1183 & 0.1359 & 0.3297 & 0.0421 & 0.3331 & 0.3264 & 0.3247 & 0.3046 & 0.0188 & 0.3111 & 0.3226 \\
 & TrainSE & 0.0049 & 0.0085 & 0.0023 & 0.0104 & 0.0022 & 0.0022 & 0.0033 & 0.0025 & 0.0002 & 0.0027 & 0.0027 \\
 & TestMSE & 0.5724 & 0.6024 & 0.5388 & 0.7313 & 0.5364 & 0.5160 & 0.5210 & 0.5344 & 1.4853 & 0.5270 & 0.9099 \\
 & TestSE & 0.0037 & 0.0056 & 0.0039 & 0.0125 & 0.0041 & 0.0038 & 0.0038 & 0.0041 & 0.0096 & 0.0039 & 0.0327 \\
\bottomrule
\end{tabular}
}
\end{table}

\FloatBarrier

\section*{KNN regression with 2-dimensional embeddings for high-dimensional, low-sample-size data}

\begin{table}[!htbp]
\centering
\caption{KNN regression with 2-dimensional embeddings for high-dimensional, low-sample-size data results for Model 1 based on datasets a1, b1, and c1.}
\label{tab:knn_2d_highdim_model1}
\resizebox{\textwidth}{!}{%
\begin{tabular}{llccccccccccc}
\toprule
Model & Type & Orig & SSU & CaSU & CoSU & UU & PCA & KPCA & SIR & KSIR & SPCA & t-SNE \\
\midrule
\multirow{4}{*}{a1} & TrainMSE & 3.7300 & 2.2528 & 3.8794 & 0.0129 & 3.8748 & 3.9247 & 3.9326 & 0.5206 & 0.1070 & 0.4534 & 3.9085 \\
 & TrainSE & 0.0415 & 0.0320 & 0.0472 & 0.0052 & 0.0428 & 0.0382 & 0.0378 & 0.0313 & 0.0027 & 0.0225 & 0.0370 \\
 & TestMSE & 7.5415 & 8.8204 & 7.7528 & 10.9987 & 7.8614 & 7.7724 & 7.7214 & 10.3470 & 11.6408 & 6.3396 & 7.8372 \\
 & TestSE & 0.0606 & 0.0834 & 0.0644 & 0.1973 & 0.0815 & 0.1068 & 0.1385 & 0.3210 & 0.0867 & 0.0520 & 0.0743 \\
\midrule
\multirow{4}{*}{b1} & TrainMSE & 2.0031 & 1.0130 & 2.0457 & 0.0302 & 2.0319 & 2.0612 & 2.0598 & 0.8801 & 0.3608 & 2.0484 & 2.0497 \\
 & TrainSE & 0.0281 & 0.0171 & 0.0238 & 0.0061 & 0.0239 & 0.0280 & 0.0240 & 0.0357 & 0.0115 & 0.0267 & 0.0262 \\
 & TestMSE & 4.0703 & 4.9172 & 4.1510 & 9.2051 & 4.1637 & 4.1923 & 4.2220 & 5.3860 & 53.6254 & 4.1896 & 52.3677 \\
 & TestSE & 0.0387 & 0.0496 & 0.0434 & 0.5829 & 0.0471 & 0.0514 & 0.0820 & 0.1533 & 0.3859 & 0.0553 & 4.6860 \\
\midrule
\multirow{4}{*}{c1} & TrainMSE & 1.2755 & 0.8888 & 1.2854 & 0.0812 & 1.2877 & 1.2758 & 1.2808 & 0.9336 & 0.2999 & 0.9874 & 1.2630 \\
 & TrainSE & 0.0178 & 0.0132 & 0.0159 & 0.0219 & 0.0175 & 0.0162 & 0.0179 & 0.0152 & 0.0087 & 0.0164 & 0.0152 \\
 & TestMSE & 2.5534 & 2.8995 & 2.5961 & 3.9842 & 2.5856 & 2.5370 & 2.5831 & 2.7229 & 17.5967 & 2.4259 & 13.6783 \\
 & TestSE & 0.0274 & 0.0299 & 0.0272 & 0.1513 & 0.0293 & 0.0285 & 0.0437 & 0.0344 & 0.1591 & 0.0285 & 0.9597 \\
\bottomrule
\end{tabular}
}
\end{table}

\begin{table}[!htbp]
\centering
\caption{KNN regression with 2-dimensional embeddings for high-dimensional, low-sample-size data results for Model 2 based on datasets a2, b2, and c2.}
\label{tab:knn_2d_highdim_model2}
\resizebox{\textwidth}{!}{%
\begin{tabular}{llccccccccccc}
\toprule
Model & Type & Orig & SSU & CaSU & CoSU & UU & PCA & KPCA & SIR & KSIR & SPCA & t-SNE \\
\midrule
\multirow{4}{*}{a2} & TrainMSE & 2.0818 & 1.2724 & 2.1125 & 0.0097 & 2.1092 & 2.1780 & 2.1313 & 0.3006 & 0.1050 & 0.2468 & 2.1184 \\
 & TrainSE & 0.0284 & 0.0193 & 0.0275 & 0.0009 & 0.0255 & 0.0295 & 0.0286 & 0.0184 & 0.0036 & 0.0055 & 0.0259 \\
 & TestMSE & 4.1322 & 4.8187 & 4.3690 & 6.1093 & 4.3343 & 4.1617 & 4.3486 & 5.1680 & 6.2938 & 3.2278 & 4.2783 \\
 & TestSE & 0.0452 & 0.0580 & 0.0556 & 0.1877 & 0.0531 & 0.0634 & 0.1333 & 0.2568 & 0.0611 & 0.0326 & 0.0465 \\
\midrule
\multirow{4}{*}{b2} & TrainMSE & 2.8199 & 1.3633 & 2.9264 & 0.0198 & 2.8948 & 2.9409 & 2.9350 & 1.1228 & 0.3907 & 2.4870 & 2.8803 \\
 & TrainSE & 0.0422 & 0.0295 & 0.0428 & 0.0018 & 0.0426 & 0.0471 & 0.0410 & 0.0580 & 0.0115 & 0.0896 & 0.0452 \\
 & TestMSE & 5.6618 & 7.1304 & 6.0344 & 10.8730 & 5.9327 & 5.9719 & 6.0872 & 7.0875 & 58.2319 & 6.2108 & 55.1329 \\
 & TestSE & 0.0693 & 0.1023 & 0.0880 & 0.4930 & 0.0924 & 0.0798 & 0.1676 & 0.2566 & 0.3728 & 0.1084 & 4.8401 \\
\midrule
\multirow{4}{*}{c2} & TrainMSE & 1.4879 & 1.0461 & 1.5268 & 0.0996 & 1.5080 & 1.4985 & 1.4965 & 1.0767 & 0.6031 & 1.1466 & 1.4825 \\
 & TrainSE & 0.0230 & 0.0192 & 0.0221 & 0.0204 & 0.0222 & 0.0198 & 0.0218 & 0.0208 & 0.0273 & 0.0231 & 0.0226 \\
 & TestMSE & 3.0234 & 3.4376 & 3.1478 & 5.2924 & 3.1100 & 2.9789 & 3.0343 & 3.2461 & 26.2139 & 2.9127 & 17.1968 \\
 & TestSE & 0.0431 & 0.0460 & 0.0413 & 0.2515 & 0.0417 & 0.0374 & 0.0441 & 0.0449 & 0.2845 & 0.0384 & 1.2620 \\
\bottomrule
\end{tabular}
}
\end{table}

\begin{table}[!htbp]
\centering
\caption{KNN regression with 2-dimensional embeddings for high-dimensional, low-sample-size data results for Model 3 based on datasets a3, b3, and c3.}
\label{tab:knn_2d_highdim_model3}
\resizebox{\textwidth}{!}{%
\begin{tabular}{llccccccccccc}
\toprule
Model & Type & Orig & SSU & CaSU & CoSU & UU & PCA & KPCA & SIR & KSIR & SPCA & t-SNE \\
\midrule
\multirow{4}{*}{a3} & TrainMSE & 0.4358 & 0.2636 & 0.4486 & 0.0010 & 0.4471 & 0.4484 & 0.4498 & 0.0562 & 0.0130 & 0.0517 & 0.4451 \\
 & TrainSE & 0.0049 & 0.0042 & 0.0044 & 0.0004 & 0.0052 & 0.0044 & 0.0047 & 0.0039 & 0.0003 & 0.0012 & 0.0050 \\
 & TestMSE & 0.8786 & 1.0226 & 0.9076 & 1.2652 & 0.8925 & 0.9085 & 0.8843 & 1.1807 & 1.3398 & 0.7078 & 0.9045 \\
 & TestSE & 0.0086 & 0.0096 & 0.0098 & 0.0231 & 0.0094 & 0.0148 & 0.0193 & 0.0468 & 0.0116 & 0.0057 & 0.0084 \\
\midrule
\multirow{4}{*}{b3} & TrainMSE & 0.4542 & 0.2612 & 0.4413 & 0.0008 & 0.4524 & 0.4522 & 0.4514 & 0.1411 & 0.0129 & 0.0546 & 0.4510 \\
 & TrainSE & 0.0055 & 0.0042 & 0.0046 & 0.0001 & 0.0051 & 0.0043 & 0.0050 & 0.0066 & 0.0003 & 0.0014 & 0.0052 \\
 & TestMSE & 0.8998 & 1.0399 & 0.9227 & 1.3382 & 0.9062 & 0.9132 & 0.9204 & 1.1122 & 1.3492 & 0.7582 & 0.9251 \\
 & TestSE & 0.0083 & 0.0104 & 0.0108 & 0.0286 & 0.0105 & 0.0107 & 0.0169 & 0.0329 & 0.0135 & 0.0074 & 0.0223 \\
\midrule
\multirow{4}{*}{c3} & TrainMSE & 0.3216 & 0.2066 & 0.3211 & 0.0201 & 0.3213 & 0.3183 & 0.3164 & 0.2512 & 0.0130 & 0.3152 & 0.3192 \\
 & TrainSE & 0.0036 & 0.0034 & 0.0035 & 0.0072 & 0.0039 & 0.0037 & 0.0034 & 0.0036 & 0.0003 & 0.0038 & 0.0034 \\
 & TestMSE & 0.6439 & 0.7272 & 0.6464 & 0.9083 & 0.6470 & 0.6407 & 0.6389 & 0.6755 & 1.4750 & 0.6383 & 1.1315 \\
 & TestSE & 0.0062 & 0.0075 & 0.0070 & 0.0189 & 0.0060 & 0.0066 & 0.0064 & 0.0083 & 0.0120 & 0.0065 & 0.0435 \\
\bottomrule
\end{tabular}
}
\end{table}